\documentclass[sigconf]{acmart}
\usepackage{amsmath}
\usepackage{bm}
\usepackage{subcaption}
\usepackage{graphicx}
\usepackage{enumitem}
\usepackage{cleveref}
\usepackage[utf8]{inputenc}
\usepackage{listings}
\usepackage{multirow}
\AtBeginDocument{%
  }

\copyrightyear{2025}
\acmYear{2025}
\setcopyright{licensedusgovmixed}\acmConference[BCB '25]{Proceedings of the 16th ACM International Conference on Bioinformatics, Computational Biology, and Health Informatics}{October 11--15, 2025}{Philadelphia, PA, USA}
\acmBooktitle{Proceedings of the 16th ACM International Conference on Bioinformatics, Computational Biology, and Health Informatics (BCB '25), October 11--15, 2025, Philadelphia, PA, USA}
\acmDOI{10.1145/3765612.3767229}
\acmISBN{979-8-4007-2200-4/2025/10}
\begin{document}

\title{Enabling Few-Shot Alzheimer's Disease Diagnosis on Biomarker Data with Tabular LLMs} 

% \author{Anonymous Authors}
\author{Sophie Kearney}
\authornote{These authors contributed equally to this research.}
\author{Shu Yang}
\authornotemark[1]
\affiliation{%
  \institution{University of Pennsylvania}
  \city{Philadelphia}
  \state{PA}
  \country{USA}
}
\email{sophie.kearney@pennmedicine.upenn.edu}
\email{yangsh@pennmedicine.upenn.edu}

\author{Zixuan Wen}
\author{Bojian Hou}
\affiliation{%
  \institution{University of Pennsylvania}
  \city{Philadelphia}
  \state{PA}
  \country{USA}
}
\email{zxwen@sas.upenn.edu}
\email{houbo@pennmedicine.upenn.edu}

\author{Duy Duong-Tran}
\affiliation{%
  \institution{United States Naval Academy}
  \city{Annapolis}
  \state{MD}
  \country{USA}
}
\email{duongtra@usna.edu}

\author{Tianlong Chen}
\affiliation{%
  \institution{University of North Carolina at Chapel Hill}
  \city{Chapel Hill}
  \state{NC}
  \country{USA}
}
\email{tianlong@cs.unc.edu}

\author{Jason Moore}
\affiliation{%
  \institution{Cedars Sinai Medical Center}
  \city{West Hollywood}
  \state{CA}
  \country{USA}
}
\email{Jason.Moore@csmc.edu}

\author{Marylyn Ritchie}
\author{Li Shen}
\authornote{Correspondence.}
\affiliation{%
  \institution{University of Pennsylvania}
  \city{Philadelphia}
  \state{PA}
  \country{USA}}
\email{marylyn@pennmedicine.upenn.edu}
\email{lishen@pennmedicine.upenn.edu}

\renewcommand{\shortauthors}{Kearney and Yang et al.}

\begin{abstract}
Early and accurate diagnosis of Alzheimer's disease (AD), a complex neurodegenerative disorder, requires analysis of heterogeneous biomarkers (e.g., neuroimaging, genetic risk factors, cognitive tests, and cerebrospinal fluid proteins) typically represented in a tabular format. With flexible few-shot reasoning, multimodal integration, and natural-language-based interpretability, large language models (LLMs) offer unprecedented opportunities for prediction with structured biomedical data. We propose a novel framework called \textbf{TAP-GPT}, \textit{\underline{T}abular \underline{A}lzheimer's \underline{P}rediction \underline{GPT}}, that adapts TableGPT2, a multimodal tabular-specialized LLM originally developed for business intelligence tasks, for AD diagnosis using structured biomarker data with small sample sizes. Our approach constructs few-shot tabular prompts using in-context learning examples from structured biomedical data and finetunes TableGPT2 using the parameter-efficient qLoRA adaption for a clinical binary classification task of AD or cognitively normal (CN). The TAP-GPT framework harnesses the powerful tabular understanding ability of TableGPT2 and the encoded prior knowledge of LLMs to outperform more advanced general-purpose LLMs and a tabular foundation model (TFM) developed for prediction tasks. To our knowledge, this is the first application of LLMs to the prediction task using tabular biomarker data, paving the way for future LLM-driven multi-agent frameworks in biomedical informatics.

\end{abstract}

\begin{CCSXML}
<ccs2012>
   <concept>
       <concept_id>10010520.10010521.10010542.10010294</concept_id>
       <concept_desc>Computer systems organization~Neural networks</concept_desc>
       <concept_significance>300</concept_significance>
       </concept>
   <concept>
       <concept_id>10010405.10010444.10010449</concept_id>
       <concept_desc>Applied computing~Health informatics</concept_desc>
       <concept_significance>500</concept_significance>
       </concept>
   <concept>
       <concept_id>10010147.10010257.10010258.10010262.10010277</concept_id>
       <concept_desc>Computing methodologies~Transfer learning</concept_desc>
       <concept_significance>300</concept_significance>
       </concept>
   <concept>
       <concept_id>10010147.10010341.10010342</concept_id>
       <concept_desc>Computing methodologies~Model development and analysis</concept_desc>
       <concept_significance>500</concept_significance>
       </concept>
   <concept>
       <concept_id>10010147.10010178.10010179</concept_id>
       <concept_desc>Computing methodologies~Natural language processing</concept_desc>
       <concept_significance>500</concept_significance>
       </concept>
 </ccs2012>
\end{CCSXML}

\ccsdesc[300]{Computer systems organization~Neural networks}
\ccsdesc[500]{Applied computing~Health informatics}
\ccsdesc[300]{Computing methodologies~Transfer learning}
\ccsdesc[500]{Computing methodologies~Model development and analysis}
\ccsdesc[500]{Computing methodologies~Natural language processing}
\keywords{Alzheimer's Disease Biomarkers, Tabular Prediction, Large Language Models, AI and Machine Learning in Healthcare}

\received{7 July 2025}
\received[revised]{10 August 2025}
\received[accepted]{27 August 2025}

\maketitle

\section{Introduction}

Alzheimer's disease (AD) is a neurodegenerative disease with a complex etiology and is the sixth leading cause of death in the United States. Globally, AD is the single most common form of dementia and accounts for over two-thirds of all 55 million dementia cases, according to World Health Organization\footnote{https://www.who.int/news-room/fact-sheets/detail/dementia}; and the projected number of AD and AD-related dementia cases could surge to more than 100 million by 2050 \cite{PRINCE201363, 2022e105}. 
AD, to date, is highly prevalent and has no curative treatments and post-treatment relapse rates remain high.
An increasing number of studies have demonstrated that early detection could result in proactive treatment and a significant slow-down in AD progression. 
Diverse AD biomarkers (e.g., neuroimaging, genetics, cognitive tests, fluid biomarkers) offer unprecedented opportunities for early diagnosis~\cite{JACK2018535, Jack539}.
For example, Quantitative templates for the progression of AD (QT-PAD)\footnote{www.pi4cs.org/qt-pad-challenge}, the AD Modeling Challenge from ADNI, gathers a list of 16 clinical, cognitive, and imaging biomarkers, such as cerebrospinal fluid (CSF) tau protein level, whole brain volume, as well as age, gender, education covariates.
However, unlike text or imaging data, such biomarker data are naturally stored in a table format with columns representing each marker and rows representing each subject. 
The table often comprises of heterogeneous data types and lacks order or locality information, which presents unique challenges for developing predictive models compared to the rich information provided in text or image data~\cite{grinsztajn2022tree,mcelfresh2023neural}.

Most existing AI models developed to handle AD biomarkers can be categorized into machine learning or classic deep neural networks with limited AD domain knowledge \cite{ShenThompson2020, Baoj2024, lin2021deep}. 
Traditional machine learning pipelines, e.g. ensembles of decision trees, have remained dominant for tabular biomedical data because they are robust to small data, whereas deep learning models often struggle to capture tabular patterns~\cite{grinsztajn2022tree,mcelfresh2023neural}. 
However, conventional models offer limited flexibility in combining modalities and do not leverage prior biomedical knowledge beyond training data. This motivates exploration of foundation models for tabular data that can perform few-shot reasoning and incorporate broad prior knowledge, potentially improving prediction from biomedical tables.
The recent advancements of Transformer-based LLMs open up new opportunities for handling tabular data \cite{fang2024large, badaro2023transformers}. 
Pretrained on extensive text corpora, LLMs exhibit emergent abilities in tabular tasks, including tabular data generation, table understanding, tabular prediction, and reasoning, \textit{etc}. \cite{zhang2023generative, wangchain, han2024large, li2024cancergpt, hegselmann2023tabllm, lee2025knowledge}. 
With prompting and finetuning, LLMs can be repurposed for tabular tasks, bridging the gap left by vanilla transformers which cannot directly input lage tables or generalize to tabular formats. 
TableGPT~\cite{zha2023tablegpt} unified tables, language, and functional commands, enabling GPT-style decoders to interpret and manipulate tables with natural language instructions.
A dedicated table encoder allows an LLM access an entire table as a compact vector, overcoming token-length limitations and capturing schema-level context. 
TableGPT2~\cite{su2024tablegpt2} scaled pretraining to 593K tables and 2.36M table-question examples, introducing a novel schema-aware table encoder with a 7B–72B parameter QWen2.5 decoder~\cite{hui2024qwen2}.
This large-scale multimodal LLM achieves remarkable accuracy gains ($\sim$35–49\% improvement over base LLMs) on several table-related business intelligence tasks, confirming the value of specialized tabular pretraining. 
Meanwhile, benchmarks like TableBench~\cite{wu2025tablebench} evaluated LLM reasoning on complex Table QA tasks and found that LLMs still perform worse than human-level on challenging multi-step reasoning with real-world tables. These studies highlight both the promise of LLMs for structured data and the remaining gap in capturing real-world tabular complexity. 

Beyond LLM-based approaches, tabular foundation models specialized for small- to medium-sized datasets have been developed. 
Notably, TabPFN~\cite{hollmann2022tabpfn, hollmann2025accurate} is a transformer model that performs in-context learning (ICL) on tabular inputs: given a small training set in a table, it instantly predicts labels for new samples without any gradient updates. 
ICL first emerged in LLMs~\cite{brown2020language} and later studies revealed that transformers can use ICL to approximate machine learning algorithms such as logistic regression and neural networks~\cite{zhou2023algorithms, muller2021transformers}.
TabPFN's single forward-pass classification, enabled by prior-fitting on 100 million synthetic datasets encoding diverse causal structures, yields impressive accuracy on par with state-of-the-art methods. In fact, TabPFN can often outperform gradient-boosted tree ensembles on small tables with many outliers or missing values, needing as little as 50\% of the data to match the accuracy of conventional models. 
Subsequent work has extended this paradigm: a retrieval-augmented finetuning strategy was proposed to adapt TabPFN to larger or more complex tasks by fetching nearest-neighbor subsets of data and finetuning the model on them~\cite{thomas2024retrieval}. This approach achieved impressive results on a 95-dataset benchmark, even outperforming tuned XGBoost models and substantially boosting TabPFN's base performance. Meanwhile, researchers have also explored lighter-weight alternatives to full model training. For example, Wu and Hou~\cite{wu2025efficient} introduced an efficient retrieval-based method that uses a frozen LLM's embedding capabilities to perform k-nearest-neighbor prediction in the LLM latent space. By avoiding any model finetuning and leveraging only the target dataset and the LLM's pretrained knowledge, their method achieved surprisingly strong tabular prediction performance. This progress underscores a growing consensus that both LLMs and specialized tabular transformers can be harnessed for table data tasks. 

However, the above approaches have been applied to heterogeneous biomedical tabular data like AD biomarkers, which pose unique challenges of multimodal feature integration and often limited sample sizes (e.g. typically only a few hundred patients in research cohorts). 
Because interpretability is critical in clinical settings, LLMs may offer an advantage by producing natural language rationales for predictions. 
We aim to fill this gap by leveraging the recent tabular-focused LLM advancements for AD diagnosis. 
In this work, we propose \textit{TAP-GPT: Tabular Alzheimer's Prediction GPT}, a novel framework that adapts the TableGPT2 multimodal LLM to the task of classifying AD vs cognitively normal (CN) individuals from biomarker tables. We formulate the problem in a tabular few-shot in-context learning paradigm: the model is prompted with a small number of example subjects with biomarkers (including features from neuroimaging, genotyping, etc.) along with their diagnoses, and asked to predict the diagnosis for a new subject, all presented in a single table. 
To adapt TableGPT2 to this new task, we apply a parameter-efficient finetuning (qLoRA)~\cite{dettmers2023qlora} using a small training set. 
The resulting TAP-GPT framework combines the powerful table comprehension and reasoning abilities of TableGPT2 with prior medical knowledge encoded in the backbone LLM's weights. Our experiments show that TAP-GPT achieves promising performance, compared to both general-purpose LLMs (which lack tabular specialization) and the tabular foundation model TabPFN on the AD classification task. 
This work represents, to our knowledge, the first application of LLMs to tabular biomarker data in biomedicine, demonstrating the potential of finetuned table-language models to tackle structured prediction with limited data in biomedical domain. 
By uniting few-shot LLM reasoning with multimodal biomarker integration, the proposed TAP-GPT paves the way for future LLM-driven multi-agent systems in AD diagnosis and general health informatics.

Our paper makes the following key contributions:
\begin{itemize}[itemsep=2pt, topsep=2pt]
    \item We propose TAP-GPT, a novel framework for few-shot tabular classification on complex, multimodal AD biomarker data that leverages the powerful language ability and encoded table-specific knowledge of LLMs through our domain-aware adaptation strategy. 
    As far as we are aware, this study represents the first use of LLMs on tabular biomarker datasets, opening new avenues for LLM-powered frameworks.
    \item We demonstrate that TAP-GPT achieves promising performance on the challenging task of few-shot tabular prediction for AD diagnosis using the public QT-PAD benchmark dataset, comparable or even outperforming more advanced generic LLMs and tabular foundation model approaches.
    \item We provide a comprehensive analysis of TAP-GPT that covers different ablations settings and the interpretability analysis of the proposed framework, enabling future development of agentic/multi-agent approaches built on tabular LLMs. 
\end{itemize}

\begin{figure*}[!t]
    \centering
    \includegraphics[width=\linewidth]{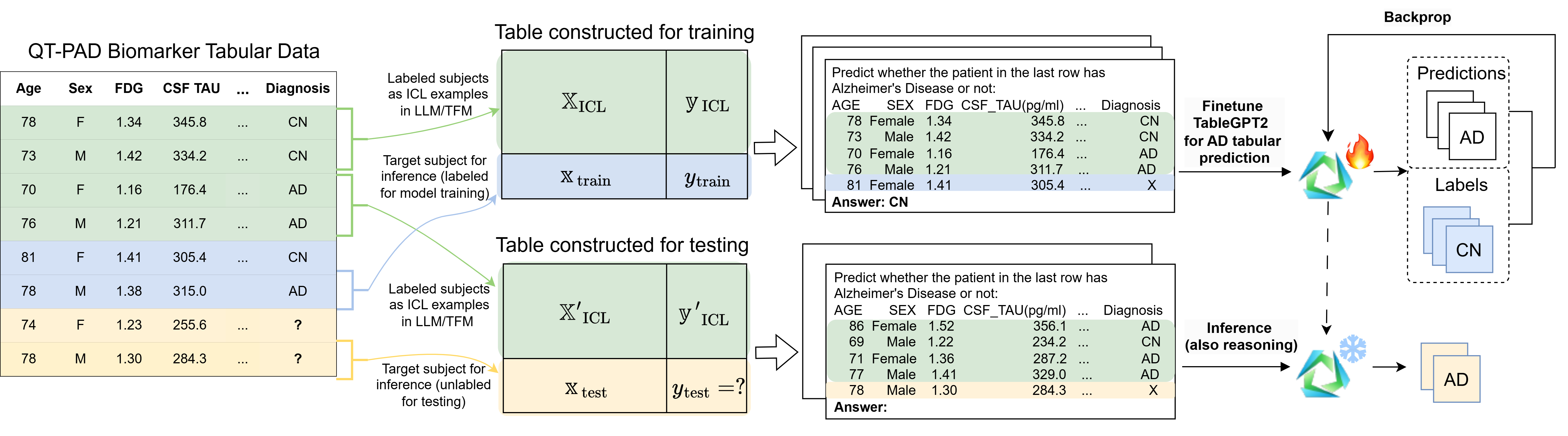}
    \caption{Overview of the TAP-GPT framework. As illustrated in the QT-PAD Biomarker Tabular Data, we split the AD and control subjects into pools for ICL (green), training (blue), and testing (yellow), from which we construct tables used in the downstream tasks (finetuning and inference).
    CN: cognitively normal. AD: Alzheimer's disease. ICL: in context learning. LLM: large language model. TFM: tabular foundation model.}
    \Description{We first extract training and testing samples from the QT-PAD biomarker table, coupling each sample with few-shot examples from the ICL pool. 
    Then, for each sample, we construct a table containing the sample and its ICL examples, and we use these tables to finetune TableGPT2 and subsequently conduct inference (or reasoning) with the resulting model.
    }
    \label{fig:overview}
\end{figure*}

\section{Methods}
We start this section by introducing our few-shot and in-context learning notions \textbf{in the context of tabular prediction tasks} following previous works~\cite{zha2023tablegpt, su2024tablegpt2, hollmann2022tabpfn, hollmann2025accurate, thomas2024retrieval, wu2025efficient}, as they are slightly different from the normal non-tabular scenario with LLMs.

Formally, let our biomarker dataset be a table $D = \{(x^i, y^i)\}_{i=1}^{N}$, consisting of $N$ labeled subject samples. 
Each sample $i$ is represented by a feature vector $x^i = (x_1^i, \dots, x_d^i) \in \mathbb{R}^d$ containing $d$ biomarkers, and a corresponding clinical diagnosis label $y^i \in \{0, 1\}$.
Here, we note 0 for CN and 1 for AD as in the QT-PAD dataset (more details in the next section).
It can be easily generalized to tabular regression or multiclass classification. 
The table columns, or features, are described by a set of natural language names for biomarkers $F = \{f_1, \dots, f_d\}$. The goal of our tabular prediction task is to learn a model that, given a new subject's feature vector $x_{test} \in \mathbb{R}^d$  and contextual examples in the form of a table, can accurately predict the corresponding label $y_{test} \in \{0, 1\}$.
An overview is provided in Figure~\ref{fig:overview}.

In the few-shot setting, the model makes this prediction conditioned on a small set of $k$ labeled examples in the context. 
This set, $D_{ICL} = \{(x_{ICL}^j, y_{ICL}^j)\}_{j=1}^k$, is provided as rows in a table to guide the model's inference.  
This tabular ICL process enables the model to approximate  
$p(y_{test} | x_{test}, D_{ICL})$ without updating its weights during inference~\cite{muller2021transformers}. In the paradigm of tabular foundation model TabPFN~\cite{hollmann2022tabpfn, hollmann2025accurate}, the training objective for a model with parameters $q_{\theta}$ is to minimize the negative log-likelihood over many such prediction tasks drawn from a training distribution $p$:
\begin{equation}
    \mathcal{L} = \mathbb{E}_{(D_{ICL}, D_{test})\sim p} [-log q_{\theta}(y_{test} | x_{test}, D_{ICL})]
\end{equation}
where $D_{test}$ is the singleton set to test containing only one element $(x_{test}, y_{test})$.

Our proposed framework, TAP-GPT, adapts this paradigm with novel modifications by leveraging the specialized architecture of TableGPT2~\cite{su2024tablegpt2}, a multimodal LLM designed for tabular data. 
To be specific, we follow the above paradigm to construct an input table $T$ of size $m \times n$, where $m=k+1$ rows and $n=d$ columns, comprising the $k$ in-context examples from $D_{ICL}$ and the single test sample $x_{test}$. 
The prediction is then generated by the model conditioned on this entire table and a task-specific instruction $I$, through standard \textbf{next token prediction} in a causal LLM:
\begin{equation}
    \hat{y}_{test} = \text{TAP-GPT}(T, I)
    \label{eqn:prediction}
\end{equation}

To predict $\hat{y}_{test}$ in Eqn~\ref{eqn:prediction}, there are methods that rely on simple serialization to convert the data $T$ into a plain text sequence and then directly prompts pretrained LLMs~\cite{hegselmann2023tabllm, lee2025knowledge}.
For example, a serialization component $\mathtt{Serialize}(\mathbf{x}^i, y^i, F)$ can take the column names $F$ and convert each row in the table to a textual description like $\text{``The} f_1 \text{ biomarker of the person is } x^i_1\text{. } ...\text{"}$; then a LLM can simply take the serialized text as input to generate the prediction, i.e. $\hat{y}_{test} = \mathtt{LLM}(\mathtt{Serialize}(\mathbf{x}^i, y^i, F))$.
However, such methods lose the structural information in the original table and can also be token-inefficient for large tables~\cite{hegselmann2023tabllm}.

Unlike these methods, our proposed TAP-GPT framework employs TableGPT2's encoder-decoder model architecture with a dedicated semantic table encoder and a LLM decoder. 
The encoder processes the input table $T$ to generate compact, structure-aware column embeddings $C(T) \in \mathbb{R}^{n \times k' \times d'}$ (where $k'$ is the number of learnable queries and $d'$ is the LLM's embedding dimension), which are then integrated with the textual instruction $I$ and processed by the LLM decoder QWen2.5~\cite{hui2024qwen2}. 
Our TAP-GPT framework is derived by supervised finetuning the decoder of the pretrained TableGPT2 on our specific AD diagnosis task, using the parameter-efficient qLoRA method~\cite{dettmers2023qlora}. 
The finetuning objective is to minimize the \textbf{cross-entropy loss} between the model's prediction $\hat{y}_{train}$ and the true label $y_{train}$ for all constructed training tables, as shown in Figure~\ref{fig:overview}.
In the following subsections, we provide more details for the data, framework and experimental setup.

\subsection{Datasets}

The data used in this study are obtained from the Alzheimer's Disease Neuroimaging Initiative (ADNI) database \footnote{\url{http://adni.loni.usc.edu}} \cite{weiner2013alzheimer,weiner2017recent}. 
 
We focus on a subset of the ADNI 1/Go/2 cohorts, i.e.,  the Quantitative templates for the progression of AD (QT-PAD) project dataset, which was initially developed as an AD Modeling Challenge from ADNI and has been widely used for diverse AD-related research. 
 
As detailed in the official website \footnote{\url{https://www.pi4cs.org/qt-pad-challenge}}, the QT-PAD dataset is processed by experts with multimodal biomarkers to describe the progression of AD, and it contains: 16 clinical markers for each subject covering PET measures (FDG PET and Amyloid PET), cerebrospinal fluid measures (CSF ABETA, CSF TAU, CSF PTau), structural MRI-derived measures from FreeSurfer (FS WholeBrain, FS Hippocampus, FS Entorhinal, FS Ventricles, FD MidTemp, FS Fusiform), and cognitive scores, as well as 1 genetic marker (APOE4 status) and 3 demographic covariates including age, gender and years of education. 
Since our aim is to predict the presence of AD in patients, we excluded the 5 cognitive markers as they are highly collinear with diagnostic outcomes and ended up with 15 features for each subject. 
 
The original cohort subset includes 699 subjects with different diagnostic groups. 
We focused on Cognitively Normal (CN) and AD subjects for our prediction task and removed samples with missing value or diagnosis. As a result, we obtained 237 CN and 96 AD subjects (333 in total).
 
All experiments are conducted on this dataset for the binary classification of CN or AD patients. 

To support ICL with tabular LLMs, we create a data splitting strategy using separate pools of data to draw ICL samples for table construction. From the 333 samples, approximately 20\% (67 samples) were allocated to the test set, 10\% (33 samples) to the validation set, and 40\% (125) to the training set. We reserve three non-overlapping pools of approximately 10\% (36 samples) for prompt construction: test set ICL pool, validation set ICL pool, and training set ICL pool. The ICL pools are never using in testing, training, or validation directly, they are only used for table construction as ICL examples. These splits are tested with ten different seeds for robustness of model evaluation. 

We designed four different prompt formats for the binary classification task to test model performance with different contexts and representations of data. Each task format involves generating a prompt with patient-level data and asking the model to predict if a patient is cognitively normal or has AD. The prompt formats vary with two dimensions: context length (zero-shot or few shot), and format (tabular or serialized). Each format is beneath a natural language prompt requesting a classification for that test sample.
 
\begin{itemize}[itemsep=2pt, topsep=2pt]
    \item \textbf{Zero-Shot Tabular} A single unlabeled row in a tabular input (one target sample).
    \item \textbf{Zero-Shot Serialized} A single unlabeled natural language description of one patient's features.
    \item \textbf{Few-Shot Tabular} For each unlabeled target sample, \textit{k} labeled rows are randomly drawn from the corresponding ICL pool as examples. These \textit{k} rows are displayed in a table above the target sample, which is the final row and is without a label. 
    \item \textbf{Few-Shot Serialized} Similar to the few-shot tabular format, but each patient's information is serialized into natural language. The \textit{k} in-context examples have labels and the one target sample does not.
\end{itemize}

Tabular prompts are displayed with the \textit{head()} function from the pandas library with the target sample in the last row with an 'X' as a label. For serialized prompts, the target sample's diagnosis sentence is omitted. Examples of each prompt format can be found in the Supplementary Materials. These four prompt formats test the impact of exposure to similar patient information (zero-shot vs. few-shot context) and data structure (serialized vs. tabular). 

\subsection{TAP-GPT framework}
TAP-GPT is a domain-adapted tabular LLM repurposed TableGPT2, which is originally optimized for business-intelligence table comprehension tasks, to a completely new structured clinical prediction task for AD.
It finetunes TableGPT2 on AD biomarker data in a structured table format with supervised labels, adapting the model from general tabular understanding to AD-specific tabular prediction.
 
Here, we evaluate if task-specific finetuning for improves performance beyond state-of-the-art methods for tabular few-shot tabular prediction while providing the reasoning capabilities of an LLM. 
As shown in Figure~\ref{fig:overview}, we first extract training and testing samples from the QT-PAD biomarker table, coupling each sample with few-shot examples from the ICL pool. 
Then, for each sample, we construct a table containing the sample and its ICL examples, and we use these tables to finetune TableGPT2 and subsequently conduct inference (or reasoning) with the resulting model.

We finetune TableGPT2-7B separately for each prompt format (zero-shot tabular, zero-shot serialized, few-shot tabular, and few-shot serialized) using the qLoRA method. qLoRA is an efficient finetuning approach using low-rank adaptation techniques that enable efficient parameter updates without modifying the full model weights. Finetuning for each task was conducted with the training set and few-shot prompts were generated with the corresponding training ICL pool. Ten random seeds (36, 73, 105, 254, 314, 492, 564, 688, 777, and 825) were used for robustness. For each task, the model was finetuned to predict the correct label, AD (1) or CN (0). We used a fixed-prompt structure across all samples with the diagnosis label always appearing in the same place after the input.

Finetuning was performed using HuggingFace's transformers and peft libraries with quantized model weights from bitsandbytes (4-bit precision) on each prompt format for each seed. During inference, a custom logits processor constrained the model's output space to only "0" or "1" for the binary classification task. 

While we finetuned TableGPT2-7B for each prompt format, the primary focus is the few-shot tabular format to augment TableGPT2 for a tabular prediction task to enable in-context tabular prediction with a limited number of examples. Comparing this setting to serialized prompts and zero-shot contexts isolates the impact of structured data and finetuning on model performance.

In this work, we only finetune TableGPT2's LLM decoder, built on Qwen2.5, keeping the tabular encoder fixed because the available pretrained TableGPT2 only exposes the decoder for downstream adaptation. While this limits the model's capacity to update tabular representations, the encoder's general-purpose table representations can be used while tailoring the LLM with domain knowledge of this clinical prediction task. 

\subsection{Experimental setup}
\begin{figure}[htbp]
  \centering
  \includegraphics[width=0.75\linewidth]{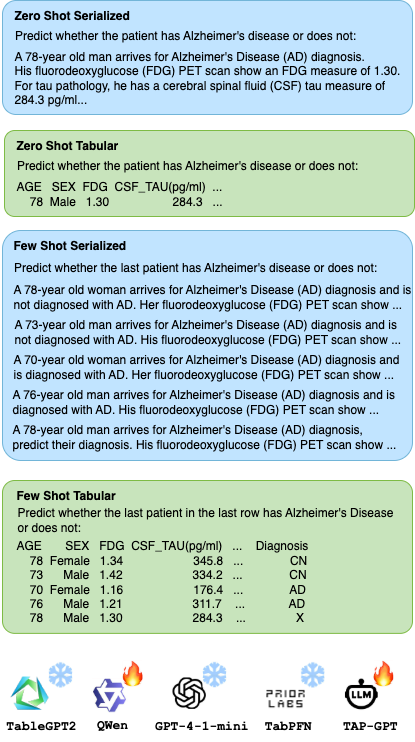}
  \caption{Prompt formats and models used in our experiments. Prompts provide subject data to LLMs, while TabPFN takes tabular CSV input. Serialized formats (blue) and tabular formats (green) are consistent across figures. Fire icons mark finetuned parameters; snowflakes mark frozen parameters.} 
  \Description{Texts of different prompt formats and icons of different LLMs/FM to illustrate the different settings in the performance experiments.}
  \label{fig:experiments}
\end{figure}

We designed a series of experiments to evaluate the performance of TAP-GPT and baseline models on a clinical classification task under varying prompt formats and model types, as shown in Figure~\ref{fig:experiments}. All experiments were repeated across ten random seeds and evaluation was performed on the test set using the four different prompt formats. The few-shot prompts were generated with \textit{k} random samples from the test ICL pool, as described in the Data section. All models were evaluated on the same binary classification task using standardized input features and the same test set. Only the prompt format and model type varied across experiments. Performance was primarily measured with F1 scores but also evaluated with accuracy, precision, and recall. 

To select an appropriate number of \textit{k} in-context examples for the few-shot prompts, we conducted a \textit{k}-ablation analysis using the validation set. TAP-GPT (finetuned TableGPT2-7B) models were trained separately on the training samples for the tabular few-shot prompt format for each $k \in {2,4,6,8,10,12,16,20}$ across the ten seeds. For each configuration, model performance was evaluated with F1 on the validation set. The \textit{k} with the highest consistent performance across all seeds on the few-shot tabular task with Finetuned TableGPT2 was chosen for all few-shot prompts.

\begin{figure*}[!t]
    \centering
    \includegraphics[width=0.8\linewidth]{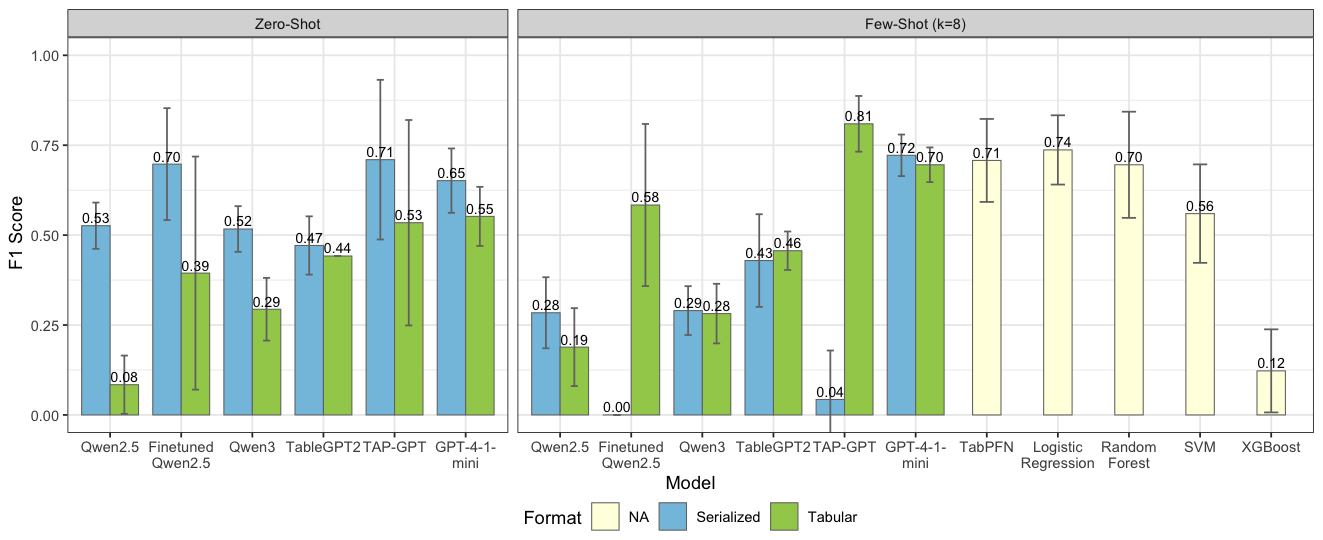}
  \Description{Bar plot showing model performance in test F1 for Alzheimer's diagnosis across zero-shot and few-shot (k=8) contexts. 
  Each LLM is tested on two prompt formats, tabular (green) and serialized (blue), with error bars indicating standard deviation. 
  In the zero-shot setting, serialized prompts outperform tabular ones, with finetuned TableGPT (TAP-GPT) achieving the highest mean F1 score. 
  In the few-shot setting, TAP-GPT performs best on tabular prompts (F1 = 0.81), followed by TabPFN and GPT-4-1-mini. 
  }
  \caption{
  Mean F1 for AD diagnosis across models in zero-shot and few-shot ($k=8$) contexts. LLMs use tabular (green) and serialized (blue) prompts with error bars for standard deviation; TabPFN and traditional ML (yellow) operate directly on structured data. TAP-GPT achieves the best performance in the tabular few-shot setting, surpassing TabPFN, general LLMs, and ML baselines.
  }
  \label{fig:F1_barplot}
\end{figure*}

To benchmark the performance of TAP-GPT, we compared model performance on all four prompt formats across a diverse set of baseline models commonly used for tabular prediction tasks and general-purpose LLMs. Unfinetuned TableGPT2 demonstrates impact of task-specific finetuning. To assess the importance of TableGPT2's tabular specialization, we tested Qwen2.5-7B-Instruct, the LLM backbone used in TableGPT2, and Qwen3-8B, a more recent general-purpose LLM from the same series. As a sanity check, we included OpenAI's GPT-4-1-mini, a significantly more capable LLM, to benchmark the upper bound of zero- and few-shot performance on this task with general-purpose LLMs. To assess the impact of finetuning TableGPT2's LLM backbone without tabular specialization, we finetuned Qwen2.5-7B-Instruct. TabPFN, a state-of-the-art tabular foundation model, was included to represent non-LLM tabular prediction and was tested only on the few-shot task with tabular data. We included four traditional machine learning models, tested only on the few-shot task: Logistic Regression as a simple baseline, Random Forest and XGBoost as tree-based ensemble methods capable of capturing nonlinear interactions, and a Support Vector Machine (SVM) for its capacity to model complex decision boundaries through kernel methods.

Hyperparameter tuning was performed using Optuna to maximize F1 on the validation set. Finetuned TableGPT2-7B and Qwen2.5 models were tuned over key parameters including LoRA rank, dropout rate, learning rate, batch size, weight decay, maximum number of steps, and type of learning rate scheduler. Traditional machine learning models were also tuned to ensure fair comparisons: Logistic Regression (C, penalty, solver, max iterations), Random Forest (number of trees, tree depth, split criteria, and sampling strategies), Support Vector Machine (C, kernel type, gamma, degree), and XGBoost (tree depth, learning rate, regularization terms, and column/row subsampling rates). The objective function maximized validation F1 and the best hyperparameter configurations were chosen for consistent good performance across the ten seeds for each model. 

All experiments were conducted on a high-performance computing cluster using a single NVIDIA A100 GPU with 80 GB of GPU memory and 160 GB of system memory. Each job was allocated 4 CPU cores and 1 node, with experiments run under a SLURM job scheduler.

\section{Results}

\subsection{Overall Model Performance}

TAP-GPT performance was evaluated on the test set across four prompt formats (zero-shot tabular, zero-shot serialized, few-shot tabular, and few-shot serialized) and compared to Qwen2.5-7B-Instruct, Finetuned Qwen2.5-7B-Instruct Qwen3.0-8B, TableGPT2, ChatGPT 4-1-mini, TabPFN, Logistic Regression, Random Forest, SVM, and XGBoost. 

As shown in Figure \ref{fig:F1_barplot} and Table \ref{table:performance_metrics}, all models performed better in the serialized than tabular setting for the zero-shot context, suggesting that LLMs struggle to extract meaningful structural information  from a single-row input. TAP-GPT achieved the highest performance in the serialized setting (mean F1 = 0.7098, SD = 0.2220), closely followed by finetuned Qwen2.5 (mean F1 = 0.6974, SD = 0.1556), suggesting that finetuning a general-purpose LLM provides sufficient domain knowledge for this task without structural cues from tabular formatting. Finetuning TableGPT yielded gains comparable to Finetuning Qwen2.5, implying that improvements in this low-context format are driven more by the LLM backbone rather than tabular prediction specialization. In the tabular zero-shot context, all models performed poorly, with GPT-4-1-mini achieving the highest score (mean F1 = 0.5521, SD = 0.0822).

In the few-shot tabular context shown in Figure \ref{fig:F1_barplot}, TAP-GPT has the strongest performance (mean F1 = 0.8095, SD = 0.0775), benefiting from pairing the tabular-aware architecture of TableGPT with domain- and task-specific knowledge from finetuning. As expected, TabPFN performs well in this context (mean F1 = 0.7077, SD = 0.1154), despite being built for larger tables. TAP-GPT performs significantly stronger than TabPFN on the tabular few-shot task (p-value=0.027). Notably, TAP-GPT also outperforms all traditional ML baselines, including logistic regression, which was the strongest among them (mean F1 = 0.737, SD = 0.0963). GPT-4-1-mini performs well across both formats in this context, which suggests that updating the backbone of TAP-GPT with more capable base models could further improve prediction performance. Most models underperform in the serialized formatting, often collapsing into single-class predictions, possibly due to prompt complexity or length. 

To assess whether TAP-GPT significantly outperforms TabPFN in the few-shot tabular setting, we evaluated differences in F1 across ten random seeds. The distribution of F1 differences satisfied the normality assumption (Shapiro-Wilk p = 0.51), supporting the use of a paired t-test, which indicated a significant improvement for TAP-GPT (p = 0.037). A Wilcoxon signed-rank test yielded a consistent result (p = 0.049), further confirming that TAP-GPT achieves significantly higher F1 scores than TabPFN. These findings reinforce the robustness of TAP-GPT’s performance advantage across evaluation metrics.

\begin{table*}[!t]
\centering
\caption{F1 and Accuracy performance metrics for each model for each prompt format. TAP-GPT (bold fonts) demonstrates the strongest performance on the Tabular Few-Shot (k=8) prompt format.}
\small 
\setlength{\tabcolsep}{3pt}
\begin{tabular}{c|c|l|r|r|r|r}
\hline
\textbf{Context} & \textbf{Format} & \textbf{Model} & \textbf{Mean F1}& \textbf{F1 SD}& \textbf{Mean Accuracy} & \textbf{SD Accuracy} \\
\hline
\multirow{14}{*}[2.7ex]{Zero-Shot}
  & \multirow{6}{*}{Serialized} & Qwen2.5 & 0.5262& 0.0643& 0.6851& 0.0514\\
  & & Finetuned Qwen2.5 & 0.6974& 0.1556& 0.8567& 0.0585\\
  & & Qwen3 & 0.5170& 0.0635& 0.6955& 0.0468\\
  & & TableGPT2 & 0.4712& 0.0810& 0.6299& 0.0667\\
  & & GPT-4-1-mini& 0.6515& 0.0895& 0.8149& 0.0446\\
  & & \textbf{TAP-GPT} & \textbf{0.7098}& \textbf{0.2220}& \textbf{0.8582}& \textbf{0.0759}\\
  \cline{2-7}
  & \multirow{8}{*}[3.0ex]{Tabular} & Qwen2.5 & 0.0840& 0.0811& 0.7149& 0.0204\\
  & & Finetuned Qwen2.5 & 0.3944& 0.3241& 0.7985& 0.0746\\
  & & Qwen3 & 0.2939& 0.0872& 0.7298& 0.0355\\
  & & TableGPT2 & 0.4419& 0.0000& 0.2836& 0.0000 \\
  & & GPT-4-1-mini & 0.5521& 0.0822& 0.7209& 0.0498\\
  & & \textbf{TAP-GPT}& \textbf{0.5345}& \textbf{0.2856}& \textbf{0.7985}& \textbf{0.0862}\\
\hline
\multirow{14}{*}[-4.5ex]{Few-Shot (k=8)}
  & \multirow{6}{*}{Serialized} & Qwen2.5 & 0.2841& 0.0988& 0.7075& 0.0374\\
  & & Finetuned Qwen2.5 & 0.0000& 0.0000 & 0.7164& 0.0000 \\
  & & Qwen3 & 0.2901& 0.0679& 0.6687& 0.0390\\
  & & TableGPT2 & 0.4293& 0.1288& 0.6567& 0.0783\\
  & & GPT-4-1-mini& 0.7219& 0.0578& 0.7910& 0.0517\\
  & & \textbf{TAP-GPT}& \textbf{0.0430}& \textbf{0.1361}&\textbf{0.6776}& \textbf{0.1227}\\
  \cline{2-7}
  & \multirow{7}{*}[1.0ex]{Tabular} & Qwen2.5 & 0.1886& 0.1083& 0.6701& 0.0395\\
  & & Finetuned Qwen2.5 & 0.5837& 0.2254& 0.8179& 0.0539\\
  & & Qwen3 & 0.2818& 0.0829& 0.6925& 0.0423\\
  & & TableGPT2 & 0.4566& 0.0535& 0.5731& 0.0493\\
  & & GPT-4-1-mini & 0.6956& 0.0482& 0.7642& 0.0481 \\
  & & \textbf{TAP-GPT} & \textbf{0.8095}& \textbf{0.0775}& \textbf{0.8880}& \textbf{0.0484}\\
  \cline{2-7}
  & \multirow{5}{*}[-0.2ex]{NA} & TabPFN & 0.7077& 0.1154& 0.8473& 0.0538\\
  & & Logistic Regression & 0.7369& 0.0963& 0.8246& 0.0732\\
  & & Random Forest & 0.6956& 0.1476& 0.8536& 0.0587\\
  & & SVM & 0.5598& 0.1369& 0.7882& 0.0542\\
  & & XGBoost & 0.1225& 0.1155& 0.6664& 0.0387\\
\hline
\end{tabular}
\label{table:performance_metrics}
\end{table*}

\subsection{Ablations}

\subsubsection{K-Ablation Analysis}

To determine the optimal $k$ in the few-shot in-context-learning setting, we conducted an ablation study on the models TAP-GPT, TableGPT2, and TabPFN for a set of different $k$s $\{2,4,6,8,10,12,16,20\}$ (Figure \ref{fig:k_ablation}).
We restricted the upper limit to 20 due to GPU memory and sample size restraints.
TabPFN showed steadily increasing performance with higher \textit{k}, consistent with its design for larger tabular input. TableGPT2 performance was relatively steady across \textit{k}, increasing after $k=2$. 
The highest mean F1 was observed at $k=10$ (0.819, SD = 0.0686), but $k=8$ (0.801, SD = 0.0644) and $k=6$ (0.799, SD = 0.0733) yielded nearly equivalent performance with overlapping variance. Because the differences among $k=6$–$10$ fall within noise and $k=8$ sits at the center of this stable region, we selected $k=8$ for all few-shot experiments.

\subsubsection{Zero-Shot Application to Few-Shot}

To evaluate if TableGPT could generalize from a simpler context, zero-shot, to a more complex context, few-shot, we tested the model finetuned on only zero-shot prompts directly on the few-shot prompts. Finetuning on a simpler task could reduce computational complexity and isolate the impact of in-context examples at the time of inference. However, the zero-shot finetuned TableGPT underperformed on the few-shot tabular prompts (mean F1 = 0.6472, SD = 0.0992) compared to TAP-GPT, TableGPT finetuned on few-shot prompts (mean F1 = 0.8095, SD = 0.0775). These results indicate that alignment of prompt formatting for finetuning and inference is critical.

\subsection{Interpretability Analysis}\label{sec:interpret}

A critical requirement for the clinical adoption of any predictive model is interpretability, the ability to understand the rationale behind decisions. 
While tabular foundation models like TabPFN demonstrate strong predictive performance, their nature as purely structural foundation models means they cannot provide innate natural language explanations for their predictions.
This \textit{black box} characteristic could be a significant barrier in biomedical domains. 
In contrast, a unique strength of our LLM-based TAP-GPT framework is its capacity to generate human-readable reasoning for its tabular predictions, like AD diagnosis here, a feature crucial for building interpretability and for future integration into trustworthy multi-agent systems (ongoing work).

To explore this capability, we conducted a case study using a minimal Chain-of-Thought (CoT) prompting~\cite{wei2022chain} to explore the reasoning ability of TAP-GPT. 
By simply modifying the instruction and appending the standard \textit{``Let's think step-by-step''} CoT string to encourage the model's response, we prompted the finetuned TAP-GPT model to output not just a 0/1 prediction, but also the reasoning process that led to it. 
This was done without providing complex reasoning examples in the prompt, allowing us to assess the model's emergent reasoning abilities after our finetuning on domain-specific, small data.

\begin{figure}[h]
  \centering
  \includegraphics[width=0.75\linewidth]{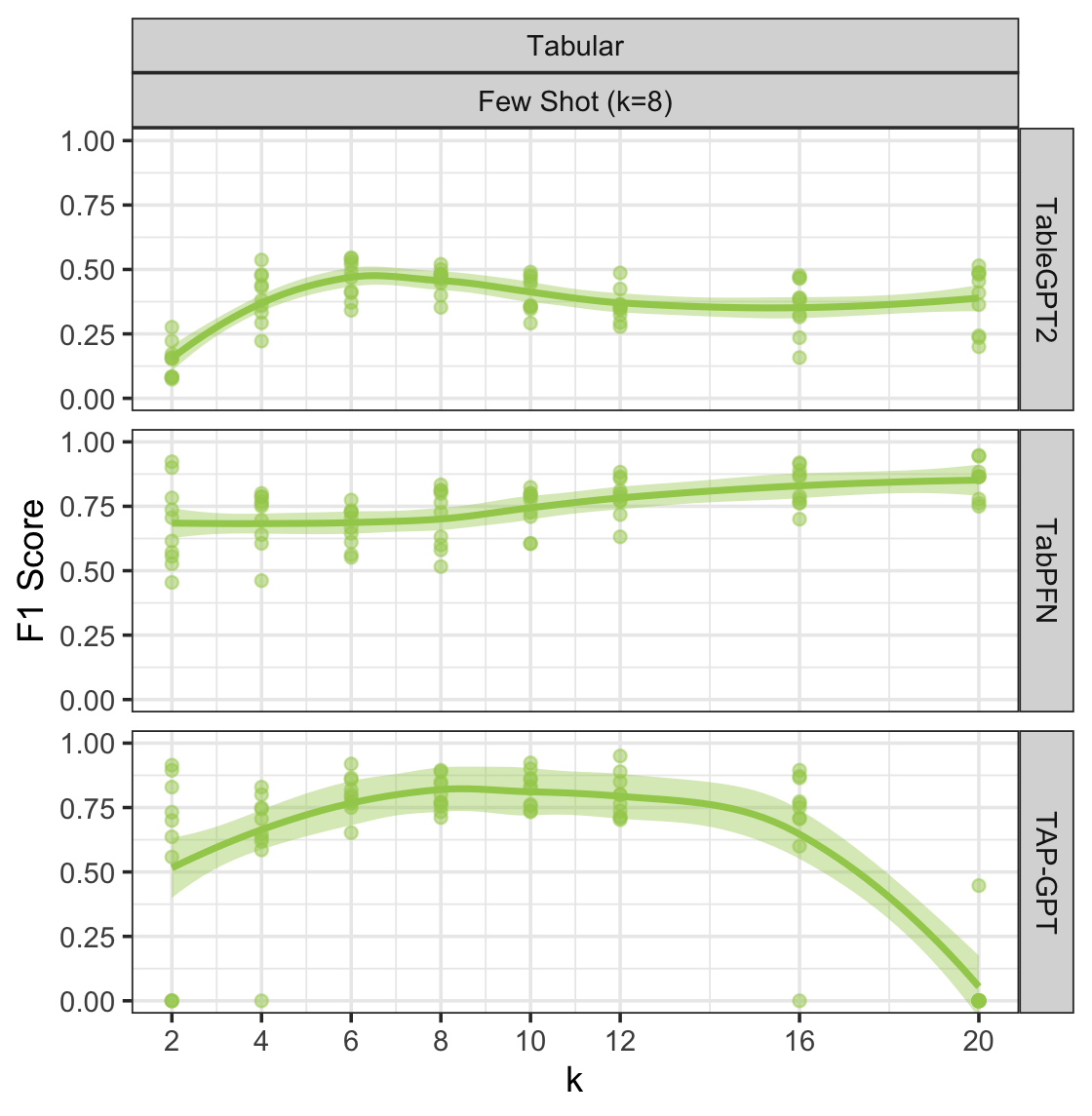}
  \Description{Grid of three line plots where the x axis is k and the y axis is F1. Ten dots are plotted for each k, indicating the F1 for each of the ten seeds. TableGPT2 shows a relatively flat line, TabPFN has a slowly increasing line, and TAP-GPT peaks around 6 to 10.}
  \caption{\textit{k} ablation analysis across TableGPT2, TabPFN, and TAP-GPT. TabPFN performance steadily improved with larger k, TableGPT2 improved up to $k=6$ and remained stable thereafter, and TAP-GPT peaked in the middle range.}
  \label{fig:k_ablation}
\end{figure}

\subsubsection*{CoT Reasoning Case Study}

We analyzed the model's generated rationales for predictions on the test set.
Two examples (manually fabricated patient data closely emulating real data) provided in the Supplementary Materials illustrate both the CoT prompt and the reasoning process to showcase TAP-GPT's interpretability.

\begin{figure}[h]
  \centering
  \includegraphics[width=0.75\linewidth]{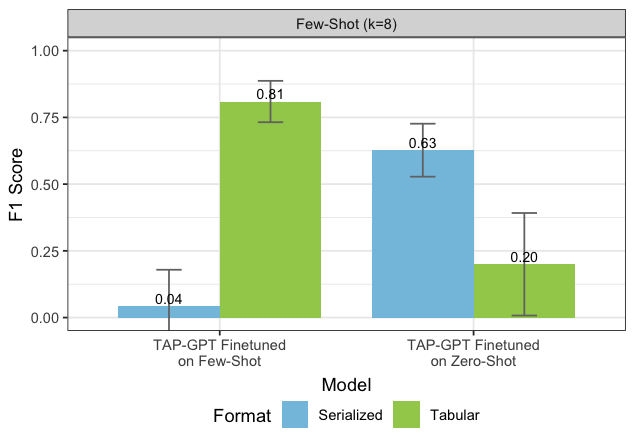}
  \Description{Barplot showing the performance of TAP-GPT and TableGPT2 finetuned for the zero-shot prompt on the few-shot prompt. TAP-GPT outperforms the zero-shot finetuned model on the tabular formatting.}
  \caption{Evaluation of TableGPT2 finetuned on zero-shot prompts used on few-shot prompts. Performance drops significantly, indicating that the finetuned model does not generalize well across prompt formats.}
  \label{fig:zeroshot_ablation}
\end{figure}

In a representative true positive case (Supplementary Materials, Example 1), TAP-GPT correctly predicted a diagnosis of AD (label 1) and generated a clinically relevant rationale. 
The model's reasoning correctly identified key risk factors from the patient's data, stating: 
 
\textit{``generally, higher values for CSF\_ABETA and lower values for CSF\_TAU and CSF\_PTAU are associated with lower risk of Alzheimer's disease... The patient is male and has APOE4, which increases the risk. The CSF\_ABETA level is low (740.4), and the CSF\_TAU and CSF\_PTAU levels are relatively high (332.0 and 31.58, respectively), indicating a higher risk... Based on the high risk factors (APOE4, low CSF\_ABETA, high CSF\_TAU and CSF\_PTAU), it is likely that this patient has Alzheimer's disease.''}
% \end{quotation} 
This explanation aligns with established AD pathology, where amyloid plaque accumulation in the brain leads to lower CSF A$\beta$ levels, and neuronal injury leads to higher CSF Tau and p-Tau levels. 
This ability to articulate a step-by-step diagnostic logic based on multiple biomarkers is a significant advantage over non-linguistic models like TabPFN.

In the second case (Supplementary Materials, Example 2), the model also made a correct prediction with step-by-step reasoning.
However, by providing the reasoning process, we can spot flaws in the model's understanding, i.e.: 
\textit{``Patients with Alzheimer's disease (AlzheimersDisease = 1) tend to have higher values for CSF\_ABETA, CSF\_TAU, and CSF\_PTAU.''} 
This statement incorrectly generalizes the elevated pattern for Tau and p-Tau to A$beta$, contradicting the known biomarker signature of AD. 
This error suggests that the model, in some cases, may oversimplify patterns from the in-context examples, adopting a naive heuristic that ``higher is worse'' for all CSF biomarkers. 
Moreover, this highlights that the reasoning process can reveal the limits of the model and shed light on potential solutions. 
Maybe the biomedical knowledge encoded in the model is incorrect, or maybe the model is too sensitive to the data distribution of the few-shot examples provided in the prompt. 
A black-box model may never reveal such limits since the final prediction is still correct.
The reasoning ability from TAP-GPT allows transparent interpretability analysis through the model's own explanation, providing important insights for further improvements, such as implementing strategies for selecting balanced and representative ICL examples, more finetuning or retrieval argumentation with AD specific materials~\cite{zhan2025general, li2024dalk} to revise the knowledge, etc.

In general, although TableGPT2 uses QWen2.5-base-7B as its backbone model, which is not a reasoning model itself, with CoT prompting fed to our finetuned TAP-GPT, we can provide detailed reasoning to the tabular prediction results for AD diagnosis. 
This is an unique strength over TabPFN and other tabular foundation models (with no language reasoning ability), offering a promising feature for future biomedical tasks and multi-agent system design

\section{Discussion}

Foundation models for text and images have revolutionized AI research and there is a growing realization that tabular data, the backbone of clinical and scientific datasets, deserves a similar focus. This work demonstrates the ability of finetuning a tabular LLM model for a clinical prediction task to outperform general-purpose LLMs, a tabular foundation model, and traditional methods on a few-shot tabular clinical prediction task. TAP-GPT benefits from the the structured setting with enough examples to reveal structural patterns across rows and columns as well as the domain-specific knowledge obtained from finetuning.

Our work pioneers the application of a tabular LLM to the challenging task of AD diagnosis based on tabular biomarker data in a data-scarce setting. 
Our proposed TAP-GPT framework successfully adapts TableGPT2 model originally designed for business intelligence into a powerful tool for few-shot AD tabular prediction. 
The results show that by combining a tabular-aware architecture with domain-specific finetuning, TAP-GPT achieves promising performance for the task. 
Our in-depth experiments demonstrate the clear synergy between the different components introduced in the TAP-GPT framework. 
As shown in Figure~\ref{fig:F1_barplot}, neither the original TableGPT2 nor the finetuned general-purpose LLM (Qwen2.5-Instruct) could match the performance of TAP-GPT in the crucial few-shot tabular setting. 
This provides insights that simply having a tabular architecture or domain knowledge alone is insufficient; their integration is likely the key for the improved capability.

An important finding of this study is TAP-GPT's ability to compete with TabPFN in the low-shot realm that is very common in biomedical research.
TabPFN, a leading tabular foundation model, represents the new generation of techniques for tabular data, an area deep learning approaches have traditionally struggled with, and introduces a paradigm shift from classic tree-based approaches. 
Our relatively simple approach performs comparatively with TabPFN in lower-shot contexts, as seen in Figure \ref{fig:k_ablation}, particularly in the 4-10 range. This suggests that TAP-GPT performs well with interpretability when labeled data are limited, a common context in the biomedical domain. At larger $k$s, the current framework of TAP-GPT may be limited by context and prompt complexity. TabPFN shows improvement over $k$, which is expected as it is a more sophisticated model built for tabular prediction tasks that can handle up to 10,000 ICL examples~\cite{hollmann2025accurate}.
However, it is worth emphasizing that, TAP-GPT, inheriting from TableGPT2, is essentially a multimodal method, connecting a tabular encoder to a large language model decoder.
Similar to vision-language models~\cite{zhang2024vision}, the table and language representations are connected in the latent embedding space through training, allowing TAP-GPT to leverage not only the structural patterns within the table but also the powerful language ability encoded in its LLM backbone. 
In contrast, TabPFN, although pretrained on millions of tables for few-shot ICL on small to medium-sized tabular data, relies solely on the tabular structure from the input data, without any language modality/ability.
Therefore, it does not have the rich semantic knowledge from textual data encoded in LLMs like Qwen that may already have prior knowledge related to AD that can help low-shots predictions, such as familiarity with the biomarker names, etc.
This may partially explain why TAP-GPT performs comparably or in some cases even better than TabPFN in the lower-shot settings.

Moreover, the potential for prediction with interpretability results through CoT reasoning provides a unique advantage of TAP-GPT over traditional machine learning models or foundation models like TabPFN. 
As demonstrated in our interpretability analysis (Section~\ref{sec:interpret}), with minimal CoT prompting, finetuned TAP-GPT can generate step-by-step rationales for its predictions. 
Revealing the reasoning that led to a particular conclusion could highlight correct and incorrect assumptions that could be adjusted to improve future predictions for more transparent decision-making in high-stakes domains like clinical prediction. It could also reveal patterns leading to a clinical prediction that have not been prioritized yet. The reasoning we observe arises from general-purpose pretraining and task-specific finetuning, so it is important to understand why the model arrives at a conclusion, whether that is domain-relevant correlations or task-aligned patterns that could be misaligned with domain knowledge. This interpretability directly supports our third contribution and lays the groundwork for more sophisticated, trustworthy AI, such as the multi-agent diagnostic systems we envision for future work.
Having said that, the CoT reasoning is an emergent capability from finetuning and can sometimes produce clinically inaccurate statements. 
We would like to continue work in this area to further understand the reasoning capabilities of TAP-GPT on tabular data to further improve our clinical predictions and provide useful insights.

Even without tabular-specific architecture, OpenAI's GPT-4-1-mini was able to achieve strong performance without finetuning. This indicates the emergent capability of current general-purpose LLMs of tabular understanding and prediction with high-capacity pretraining. Future directions for this framework could include manually decoupling the architecture of TableGPT2 to replace the current LLM decoder, Qwen2.5, with a more modern LLM that already has some tabular understanding capacities. However, the current architecture of TableGPT2 is not open-source, so this limits our ability to modify parts of its structure. Manually recreating TableGPT's architecture, potentially using TabPFN's tabular encoder and a general-purpose LLM like OpenAI's GPT-4-1-mini or Meta's Llama-3-8, could be a future step to further specialize a tabular LLM for clinical prediction tasks.

Our current dataset contains information relating to the ATN (Amyloid, Tau, and Neurodegeneration) framework of AD diagnosis, but we would like to expand to incorporate additional biomarkers like genetic data or imaging data to create a more robust multi-modal analysis. Multimodal integration could improve prediction performance but also reveal latent interactions in disease progression and prediction. Systematic ablation or attribution-based analyses could help identify which data types drive model predictions in different contexts, aiding interpretability of results. Additionally, expanding to more difficult tasks like future prediction of AD or differentiation between cognitively normal, mild cognitive impairment, and AD could further distinguish the capabilities of TAP-GPT.
Moreover, since the current study is based on QT-PAD with a moderate sample size, our future goal is to extend the study to larger, more diverse cohorts, and for different diseases so that can evaluate more generalized performances of TAP-GPT.

One promising advantage of LLM-based prediction models is their potential for handling missingness. Traditional machine learning methods or TFMs require imputation if missing values occur, which can mislead prediction. Because LLMs operate over text-like serialized inputs, missingness can be implicitly represented by the absence of a value, rather than forcing the model to reason over artificially filled-in data. The LLM could learn context-dependent patterns of missingness and use it to infer as an informative signal, which could further enhance prediction. Accurate handling of missingness is critical in clinical tasks, where data is frequently incomplete or irregular and accurate diagnosis is critical. This is a future area of analysis for TAP-GPT and other general-purpose LLMs on this task.

Lastly, this work lays the foundation for developing LLM-driven multi-agent AD diagnostic systems. 
We envision a framework where TAP-GPT acts as a specialized biomarker agent, generating some initial diagnosis and rationale. 
This output could then be passed to another independent neuroimaging agent, or an EHR agent etc., which could leverage additional information and knowledge to cross-validate the clinical reasoning. 
Such a collaborative system could mitigate the risk of flawed individual judgments and represents a potential step towards building more reliable and trustworthy AI in AD diagnosis and other fields.
This work serves as a successful proof-of-concept and provides the foundation for adapting emerging tabular LLMs to solve these critical problems.

\section{Conclusion}
This work demonstrates the promising strength of TAP-GPT on tabular few-shot prompts for predicting Alzheimer's Disease in patients using clinical biomarkers from the ATN framework. TAP-GPT expands the capabilities of TableGPT2 beyond tabular understanding into tabular prediction. TAP-GPT outperforms TabPFN, a tabular foundation model, and general-purpose LLMs in this low-shot context, leveraging structural information and task-specific finetuning. Beyond performance, TAP-GPT offers interpretability in its reasoning for predictions, which could aid clinical implementation and offer use in multi-agent frameworks. Our results indicate the importance of prompt context (zero-shot vs few shot, tabular vs. serialized) on LLM prediction performance. Our framework serves as a step to apply LLMs to tabular biomarker data to support future clinical prediction tasks and incorporation into multi-agent frameworks.

\begin{acks}
This work was supported in part by the NIH grants U01 AG066833, P30 AG073105, U01 AG068057, and U19 AG074879.

Data collection and sharing for this project was funded by the Alzheimer's Disease Neuroimaging Initiative (ADNI) (National Institutes of Health Grant U01 AG024904) and DOD ADNI (Department of Defense award number W81XWH-12-2-0012).

\end{acks}

\section*{Disclaimer}
The views expressed in this manuscript are those of the author(s) and do not reflect the official policy or position of the U.S. Naval Academy, Department of the Navy, the Department of Defense, or the U.S.Government.

\bibliographystyle{ACM-Reference-Format}
\bibliography{main}


\begin{thebibliography}{35}

%%% the Web address.

\ifx \showCODEN    \undefined \def \showCODEN     #1{\unskip}     \fi
\ifx \showISBNx    \undefined \def \showISBNx     #1{\unskip}     \fi
\ifx \showISBNxiii \undefined \def \showISBNxiii  #1{\unskip}     \fi
\ifx \showISSN     \undefined \def \showISSN      #1{\unskip}     \fi
\ifx \showLCCN     \undefined \def \showLCCN      #1{\unskip}     \fi
\ifx \shownote     \undefined \def \shownote      #1{#1}          \fi
\ifx \showarticletitle \undefined \def \showarticletitle #1{#1}   \fi
\ifx \showURL      \undefined \def \showURL       {\relax}        \fi
% invisible to TeX
\providecommand\bibfield[2]{#2}
\providecommand\bibinfo[2]{#2}
\providecommand\natexlab[1]{#1}
\providecommand\showeprint[2][]{arXiv:#2}

\bibitem[Badaro et~al\mbox{.}(2023)]%
        {badaro2023transformers}
\bibfield{author}{\bibinfo{person}{Gilbert Badaro}, \bibinfo{person}{Mohammed Saeed}, {and} \bibinfo{person}{Paolo Papotti}.} \bibinfo{year}{2023}\natexlab{}.
\newblock \showarticletitle{Transformers for Tabular Data Representation: A Survey of Models and Applications}.
\newblock \bibinfo{journal}{\emph{Transactions of the Association for Computational Linguistics}}  \bibinfo{volume}{11} (\bibinfo{year}{2023}), \bibinfo{pages}{227--249}.
\newblock


\bibitem[Bao et~al\mbox{.}(2024)]%
        {Baoj2024}
\bibfield{author}{\bibinfo{person}{Jingxuan Bao}, \bibinfo{person}{Brian~N. Lee}, \bibinfo{person}{Junhao Wen}, \bibinfo{person}{Mansu Kim}, \bibinfo{person}{Shizhuo Mu}, \bibinfo{person}{Shu Yang}, \bibinfo{person}{Christos Davatzikos}, \bibinfo{person}{Qi Long}, \bibinfo{person}{Marylyn~D. Ritchie}, {and} \bibinfo{person}{Li Shen}.} \bibinfo{year}{2024}\natexlab{}.
\newblock \showarticletitle{Employing Informatics Strategies in Alzheimer Disease Research: A Review from Genetics, Multiomics, and Biomarkers to Clinical Outcomes}.
\newblock \bibinfo{journal}{\emph{Annual Review of Biomedical Data Science}} (\bibinfo{year}{2024}).
\newblock
\href{https://doi.org/10.1146/annurev-biodatasci-102423-121021}{doi:\nolinkurl{10.1146/annurev-biodatasci-102423-121021}}


\bibitem[Brown et~al\mbox{.}(2020)]%
        {brown2020language}
\bibfield{author}{\bibinfo{person}{Tom Brown}, \bibinfo{person}{Benjamin Mann}, \bibinfo{person}{Nick Ryder}, \bibinfo{person}{Melanie Subbiah}, \bibinfo{person}{Jared~D Kaplan}, \bibinfo{person}{Prafulla Dhariwal}, \bibinfo{person}{Arvind Neelakantan}, \bibinfo{person}{Pranav Shyam}, \bibinfo{person}{Girish Sastry}, \bibinfo{person}{Amanda Askell}, {et~al\mbox{.}}} \bibinfo{year}{2020}\natexlab{}.
\newblock \showarticletitle{Language models are few-shot learners}.
\newblock \bibinfo{journal}{\emph{Advances in neural information processing systems}}  \bibinfo{volume}{33} (\bibinfo{year}{2020}), \bibinfo{pages}{1877--1901}.
\newblock


\bibitem[Dettmers et~al\mbox{.}(2023)]%
        {dettmers2023qlora}
\bibfield{author}{\bibinfo{person}{Tim Dettmers}, \bibinfo{person}{Artidoro Pagnoni}, \bibinfo{person}{Ari Holtzman}, {and} \bibinfo{person}{Luke Zettlemoyer}.} \bibinfo{year}{2023}\natexlab{}.
\newblock \showarticletitle{Qlora: Efficient finetuning of quantized llms}.
\newblock \bibinfo{journal}{\emph{Advances in neural information processing systems}}  \bibinfo{volume}{36} (\bibinfo{year}{2023}), \bibinfo{pages}{10088--10115}.
\newblock


\bibitem[Fang et~al\mbox{.}(2024)]%
        {fang2024large}
\bibfield{author}{\bibinfo{person}{Xi Fang}, \bibinfo{person}{Weijie Xu}, \bibinfo{person}{Fiona~Anting Tan}, \bibinfo{person}{Ziqing Hu}, \bibinfo{person}{Jiani Zhang}, \bibinfo{person}{Yanjun Qi}, \bibinfo{person}{Srinivasan~H. Sengamedu}, {and} \bibinfo{person}{Christos Faloutsos}.} \bibinfo{year}{2024}\natexlab{}.
\newblock \showarticletitle{Large Language Models ({LLM}s) on Tabular Data: Prediction, Generation, and Understanding - A Survey}.
\newblock \bibinfo{journal}{\emph{Transactions on Machine Learning Research}} (\bibinfo{year}{2024}).
\newblock
\showISSN{2835-8856}


\bibitem[Grinsztajn et~al\mbox{.}(2022)]%
        {grinsztajn2022tree}
\bibfield{author}{\bibinfo{person}{L{\'e}o Grinsztajn}, \bibinfo{person}{Edouard Oyallon}, {and} \bibinfo{person}{Ga{\"e}l Varoquaux}.} \bibinfo{year}{2022}\natexlab{}.
\newblock \showarticletitle{Why do tree-based models still outperform deep learning on typical tabular data?}
\newblock \bibinfo{journal}{\emph{Advances in neural information processing systems}}  \bibinfo{volume}{35} (\bibinfo{year}{2022}), \bibinfo{pages}{507--520}.
\newblock


\bibitem[Han et~al\mbox{.}(2024)]%
        {han2024large}
\bibfield{author}{\bibinfo{person}{Sungwon Han}, \bibinfo{person}{Jinsung Yoon}, \bibinfo{person}{Sercan~O Arik}, {and} \bibinfo{person}{Tomas Pfister}.} \bibinfo{year}{2024}\natexlab{}.
\newblock \showarticletitle{Large Language Models Can Automatically Engineer Features for Few-Shot Tabular Learning}. In \bibinfo{booktitle}{\emph{Forty-first International Conference on Machine Learning}}.
\newblock


\bibitem[Hegselmann et~al\mbox{.}(2023)]%
        {hegselmann2023tabllm}
\bibfield{author}{\bibinfo{person}{Stefan Hegselmann}, \bibinfo{person}{Alejandro Buendia}, \bibinfo{person}{Hunter Lang}, \bibinfo{person}{Monica Agrawal}, \bibinfo{person}{Xiaoyi Jiang}, {and} \bibinfo{person}{David Sontag}.} \bibinfo{year}{2023}\natexlab{}.
\newblock \showarticletitle{Tabllm: Few-shot classification of tabular data with large language models}. In \bibinfo{booktitle}{\emph{International Conference on Artificial Intelligence and Statistics}}. PMLR, \bibinfo{pages}{5549--5581}.
\newblock


\bibitem[Hollmann et~al\mbox{.}(2022)]%
        {hollmann2022tabpfn}
\bibfield{author}{\bibinfo{person}{Noah Hollmann}, \bibinfo{person}{Samuel M{\"u}ller}, \bibinfo{person}{Katharina Eggensperger}, {and} \bibinfo{person}{Frank Hutter}.} \bibinfo{year}{2022}\natexlab{}.
\newblock \showarticletitle{Tabpfn: A transformer that solves small tabular classification problems in a second}.
\newblock \bibinfo{journal}{\emph{arXiv preprint arXiv:2207.01848}} (\bibinfo{year}{2022}).
\newblock


\bibitem[Hollmann et~al\mbox{.}(2025)]%
        {hollmann2025accurate}
\bibfield{author}{\bibinfo{person}{Noah Hollmann}, \bibinfo{person}{Samuel M{\"u}ller}, \bibinfo{person}{Lennart Purucker}, \bibinfo{person}{Arjun Krishnakumar}, \bibinfo{person}{Max K{\"o}rfer}, \bibinfo{person}{Shi~Bin Hoo}, \bibinfo{person}{Robin~Tibor Schirrmeister}, {and} \bibinfo{person}{Frank Hutter}.} \bibinfo{year}{2025}\natexlab{}.
\newblock \showarticletitle{Accurate predictions on small data with a tabular foundation model}.
\newblock \bibinfo{journal}{\emph{Nature}} \bibinfo{volume}{637}, \bibinfo{number}{8045} (\bibinfo{year}{2025}), \bibinfo{pages}{319--326}.
\newblock


\bibitem[Hui et~al\mbox{.}(2024)]%
        {hui2024qwen2}
\bibfield{author}{\bibinfo{person}{Binyuan Hui}, \bibinfo{person}{Jian Yang}, \bibinfo{person}{Zeyu Cui}, \bibinfo{person}{Jiaxi Yang}, \bibinfo{person}{Dayiheng Liu}, \bibinfo{person}{Lei Zhang}, \bibinfo{person}{Tianyu Liu}, \bibinfo{person}{Jiajun Zhang}, \bibinfo{person}{Bowen Yu}, \bibinfo{person}{Keming Lu}, {et~al\mbox{.}}} \bibinfo{year}{2024}\natexlab{}.
\newblock \showarticletitle{Qwen2. 5-coder technical report}.
\newblock \bibinfo{journal}{\emph{arXiv preprint arXiv:2409.12186}} (\bibinfo{year}{2024}).
\newblock


\bibitem[Jack et~al\mbox{.}(2016)]%
        {Jack539}
\bibfield{author}{\bibinfo{person}{C.~R. Jack}, \bibinfo{person}{D.~A. Bennett}, {et~al\mbox{.}}} \bibinfo{year}{2016}\natexlab{}.
\newblock \showarticletitle{A/T/N: An unbiased descriptive classification scheme for Alzheimer disease biomarkers}.
\newblock \bibinfo{journal}{\emph{Neurology}} \bibinfo{volume}{87}, \bibinfo{number}{5} (\bibinfo{year}{2016}), \bibinfo{pages}{539--547}.
\newblock
\showISSN{0028-3878}
\href{https://doi.org/10.1212/WNL.0000000000002923}{doi:\nolinkurl{10.1212/WNL.0000000000002923}}


\bibitem[Jack et~al\mbox{.}(2018)]%
        {JACK2018535}
\bibfield{author}{\bibinfo{person}{C.~R. Jack}, \bibinfo{person}{D.~A. Bennett}, {et~al\mbox{.}}} \bibinfo{year}{2018}\natexlab{}.
\newblock \showarticletitle{NIA-AA Research Framework: Toward a biological definition of Alzheimer's disease}.
\newblock \bibinfo{journal}{\emph{Alzheimer's and Dementia}} \bibinfo{volume}{14}, \bibinfo{number}{4} (\bibinfo{year}{2018}), \bibinfo{pages}{535--562}.
\newblock
\showISSN{1552-5260}
\href{https://doi.org/10.1016/j.jalz.2018.02.018}{doi:\nolinkurl{10.1016/j.jalz.2018.02.018}}


\bibitem[Lee et~al\mbox{.}(2025)]%
        {lee2025knowledge}
\bibfield{author}{\bibinfo{person}{Joseph Lee}, \bibinfo{person}{Shu Yang}, \bibinfo{person}{Jae~Young Baik}, \bibinfo{person}{Xiaoxi Liu}, \bibinfo{person}{Zhen Tan}, \bibinfo{person}{Dawei Li}, \bibinfo{person}{Zixuan Wen}, \bibinfo{person}{Bojian Hou}, \bibinfo{person}{Duy Duong-Tran}, \bibinfo{person}{Tianlong Chen}, {et~al\mbox{.}}} \bibinfo{year}{2025}\natexlab{}.
\newblock \showarticletitle{Knowledge-driven feature selection and engineering for genotype data with large language models}.
\newblock \bibinfo{journal}{\emph{AMIA Summits on Translational Science Proceedings}}  \bibinfo{volume}{2025} (\bibinfo{year}{2025}), \bibinfo{pages}{250}.
\newblock


\bibitem[Li et~al\mbox{.}(2024b)]%
        {li2024dalk}
\bibfield{author}{\bibinfo{person}{Dawei Li}, \bibinfo{person}{Shu Yang}, \bibinfo{person}{Zhen Tan}, \bibinfo{person}{Jae~Young Baik}, \bibinfo{person}{Sunkwon Yun}, \bibinfo{person}{Joseph Lee}, \bibinfo{person}{Aaron Chacko}, \bibinfo{person}{Bojian Hou}, \bibinfo{person}{Duy Duong-Tran}, \bibinfo{person}{Ying Ding}, {et~al\mbox{.}}} \bibinfo{year}{2024}\natexlab{b}.
\newblock \showarticletitle{DALK: Dynamic Co-Augmentation of LLMs and KG to answer Alzheimer's Disease Questions with Scientific Literature}.
\newblock \bibinfo{journal}{\emph{arXiv preprint arXiv:2405.04819}} (\bibinfo{year}{2024}).
\newblock


\bibitem[Li et~al\mbox{.}(2024a)]%
        {li2024cancergpt}
\bibfield{author}{\bibinfo{person}{Tianhao Li}, \bibinfo{person}{Sandesh Shetty}, \bibinfo{person}{Advaith Kamath}, \bibinfo{person}{Ajay Jaiswal}, \bibinfo{person}{Xiaoqian Jiang}, \bibinfo{person}{Ying Ding}, {and} \bibinfo{person}{Yejin Kim}.} \bibinfo{year}{2024}\natexlab{a}.
\newblock \showarticletitle{CancerGPT for few shot drug pair synergy prediction using large pretrained language models}.
\newblock \bibinfo{journal}{\emph{NPJ Digital Medicine}} \bibinfo{volume}{7}, \bibinfo{number}{1} (\bibinfo{year}{2024}), \bibinfo{pages}{40}.
\newblock


\bibitem[Lin et~al\mbox{.}(2021)]%
        {lin2021deep}
\bibfield{author}{\bibinfo{person}{Eugene Lin}, \bibinfo{person}{Chieh-Hsin Lin}, {and} \bibinfo{person}{Hsien-Yuan Lane}.} \bibinfo{year}{2021}\natexlab{}.
\newblock \showarticletitle{Deep learning with neuroimaging and genomics in Alzheimer’s disease}.
\newblock \bibinfo{journal}{\emph{International journal of molecular sciences}} \bibinfo{volume}{22}, \bibinfo{number}{15} (\bibinfo{year}{2021}), \bibinfo{pages}{7911}.
\newblock


\bibitem[McElfresh et~al\mbox{.}(2024)]%
        {mcelfresh2023neural}
\bibfield{author}{\bibinfo{person}{Duncan McElfresh}, \bibinfo{person}{Sujay Khandagale}, \bibinfo{person}{Jonathan Valverde}, \bibinfo{person}{Vishak Prasad~C}, \bibinfo{person}{Ganesh Ramakrishnan}, \bibinfo{person}{Micah Goldblum}, {and} \bibinfo{person}{Colin White}.} \bibinfo{year}{2024}\natexlab{}.
\newblock \showarticletitle{When do neural nets outperform boosted trees on tabular data?}
\newblock \bibinfo{journal}{\emph{Advances in Neural Information Processing Systems}}  \bibinfo{volume}{36} (\bibinfo{year}{2024}), \bibinfo{pages}{76336--76369}.
\newblock


\bibitem[M{\"u}ller et~al\mbox{.}(2021)]%
        {muller2021transformers}
\bibfield{author}{\bibinfo{person}{Samuel M{\"u}ller}, \bibinfo{person}{Noah Hollmann}, \bibinfo{person}{Sebastian~Pineda Arango}, \bibinfo{person}{Josif Grabocka}, {and} \bibinfo{person}{Frank Hutter}.} \bibinfo{year}{2021}\natexlab{}.
\newblock \showarticletitle{Transformers can do bayesian inference}.
\newblock \bibinfo{journal}{\emph{arXiv preprint arXiv:2112.10510}} (\bibinfo{year}{2021}).
\newblock


\bibitem[Nichols et~al\mbox{.}(2022)]%
        {2022e105}
\bibfield{author}{\bibinfo{person}{Emma Nichols} {et~al\mbox{.}}} \bibinfo{year}{2022}\natexlab{}.
\newblock \showarticletitle{Estimation of the global prevalence of dementia in 2019 and forecasted prevalence in 2050: an analysis for the Global Burden of Disease Study 2019}.
\newblock \bibinfo{journal}{\emph{The Lancet Public Health}} \bibinfo{volume}{7}, \bibinfo{number}{2} (\bibinfo{year}{2022}), \bibinfo{pages}{e105--e125}.
\newblock
\showISSN{2468-2667}


\bibitem[Prince et~al\mbox{.}(2013)]%
        {PRINCE201363}
\bibfield{author}{\bibinfo{person}{Martin Prince}, \bibinfo{person}{Renata Bryce}, \bibinfo{person}{Emiliano Albanese}, \bibinfo{person}{Anders Wimo}, \bibinfo{person}{Wagner Ribeiro}, {and} \bibinfo{person}{Cleusa~P. Ferri}.} \bibinfo{year}{2013}\natexlab{}.
\newblock \showarticletitle{The global prevalence of dementia: A systematic review and metaanalysis}.
\newblock \bibinfo{journal}{\emph{Alzheimer's \& Dementia}} \bibinfo{volume}{9}, \bibinfo{number}{1} (\bibinfo{year}{2013}), \bibinfo{pages}{63--75.e2}.
\newblock
\showISSN{1552-5260}


\bibitem[Shen and Thompson(2020)]%
        {ShenThompson2020}
\bibfield{author}{\bibinfo{person}{Li Shen} {and} \bibinfo{person}{Paul~M. Thompson}.} \bibinfo{year}{2020}\natexlab{}.
\newblock \showarticletitle{Brain Imaging Genomics: Integrated Analysis and Machine Learning}.
\newblock \bibinfo{journal}{\emph{Proc. IEEE}} \bibinfo{volume}{108}, \bibinfo{number}{1} (\bibinfo{year}{2020}), \bibinfo{pages}{125--162}.
\newblock
\href{https://doi.org/10.1109/JPROC.2019.2947272}{doi:\nolinkurl{10.1109/JPROC.2019.2947272}}


\bibitem[Su et~al\mbox{.}(2024)]%
        {su2024tablegpt2}
\bibfield{author}{\bibinfo{person}{Aofeng Su}, \bibinfo{person}{Aowen Wang}, \bibinfo{person}{Chao Ye}, \bibinfo{person}{Chen Zhou}, \bibinfo{person}{Ga Zhang}, \bibinfo{person}{Gang Chen}, \bibinfo{person}{Guangcheng Zhu}, \bibinfo{person}{Haobo Wang}, \bibinfo{person}{Haokai Xu}, \bibinfo{person}{Hao Chen}, {et~al\mbox{.}}} \bibinfo{year}{2024}\natexlab{}.
\newblock \showarticletitle{Tablegpt2: A large multimodal model with tabular data integration}.
\newblock \bibinfo{journal}{\emph{arXiv preprint arXiv:2411.02059}} (\bibinfo{year}{2024}).
\newblock


\bibitem[Thomas et~al\mbox{.}(2024)]%
        {thomas2024retrieval}
\bibfield{author}{\bibinfo{person}{Valentin Thomas}, \bibinfo{person}{Junwei Ma}, \bibinfo{person}{Rasa Hosseinzadeh}, \bibinfo{person}{Keyvan Golestan}, \bibinfo{person}{Guangwei Yu}, \bibinfo{person}{Maks Volkovs}, {and} \bibinfo{person}{Anthony~L Caterini}.} \bibinfo{year}{2024}\natexlab{}.
\newblock \showarticletitle{Retrieval \& fine-tuning for in-context tabular models}.
\newblock \bibinfo{journal}{\emph{Advances in Neural Information Processing Systems}}  \bibinfo{volume}{37} (\bibinfo{year}{2024}), \bibinfo{pages}{108439--108467}.
\newblock


\bibitem[Wang et~al\mbox{.}(2024)]%
        {wangchain}
\bibfield{author}{\bibinfo{person}{Zilong Wang}, \bibinfo{person}{Hao Zhang}, \bibinfo{person}{Chun-Liang Li}, \bibinfo{person}{Julian~Martin Eisenschlos}, \bibinfo{person}{Vincent Perot}, \bibinfo{person}{Zifeng Wang}, \bibinfo{person}{Lesly Miculicich}, \bibinfo{person}{Yasuhisa Fujii}, \bibinfo{person}{Jingbo Shang}, \bibinfo{person}{Chen-Yu Lee}, {et~al\mbox{.}}} \bibinfo{year}{2024}\natexlab{}.
\newblock \showarticletitle{Chain-of-Table: Evolving Tables in the Reasoning Chain for Table Understanding}. In \bibinfo{booktitle}{\emph{The Twelfth International Conference on Learning Representations}}.
\newblock


\bibitem[Wei et~al\mbox{.}(2022)]%
        {wei2022chain}
\bibfield{author}{\bibinfo{person}{Jason Wei}, \bibinfo{person}{Xuezhi Wang}, \bibinfo{person}{Dale Schuurmans}, \bibinfo{person}{Maarten Bosma}, \bibinfo{person}{Fei Xia}, \bibinfo{person}{Ed Chi}, \bibinfo{person}{Quoc~V Le}, \bibinfo{person}{Denny Zhou}, {et~al\mbox{.}}} \bibinfo{year}{2022}\natexlab{}.
\newblock \showarticletitle{Chain-of-thought prompting elicits reasoning in large language models}.
\newblock \bibinfo{journal}{\emph{Advances in neural information processing systems}}  \bibinfo{volume}{35} (\bibinfo{year}{2022}), \bibinfo{pages}{24824--24837}.
\newblock


\bibitem[Weiner et~al\mbox{.}(2013)]%
        {weiner2013alzheimer}
\bibfield{author}{\bibinfo{person}{Michael~W Weiner}, \bibinfo{person}{Dallas~P Veitch}, \bibinfo{person}{Paul~S Aisen}, \bibinfo{person}{Laurel~A Beckett}, \bibinfo{person}{Nigel~J Cairns}, \bibinfo{person}{Robert~C Green}, \bibinfo{person}{Danielle Harvey}, \bibinfo{person}{Clifford~R Jack}, \bibinfo{person}{William Jagust}, \bibinfo{person}{Enchi Liu}, {et~al\mbox{.}}} \bibinfo{year}{2013}\natexlab{}.
\newblock \showarticletitle{The Alzheimer's Disease Neuroimaging Initiative: a review of papers published since its inception}.
\newblock \bibinfo{journal}{\emph{Alzheimer's \& Dementia}} \bibinfo{volume}{9}, \bibinfo{number}{5} (\bibinfo{year}{2013}), \bibinfo{pages}{e111--e194}.
\newblock


\bibitem[Weiner et~al\mbox{.}(2017)]%
        {weiner2017recent}
\bibfield{author}{\bibinfo{person}{Michael~W Weiner}, \bibinfo{person}{klas~P Veitch}, \bibinfo{person}{Paul~S Aisen}, \bibinfo{person}{Laurel~A Beckett}, \bibinfo{person}{Nigel~J Cairns}, \bibinfo{person}{Robert~C Green}, \bibinfo{person}{Danielle Harvey}, \bibinfo{person}{Clifford~R Jack~Jr}, \bibinfo{person}{William Jagust}, \bibinfo{person}{John~C Morris}, {et~al\mbox{.}}} \bibinfo{year}{2017}\natexlab{}.
\newblock \showarticletitle{Recent publications from the Alzheimer's Disease Neuroimaging Initiative: Reviewing progress toward improved AD clinical trials}.
\newblock \bibinfo{journal}{\emph{Alzheimer's \& Dementia}} \bibinfo{volume}{13}, \bibinfo{number}{4} (\bibinfo{year}{2017}), \bibinfo{pages}{e1--e85}.
\newblock


\bibitem[Wu and Hou(2025)]%
        {wu2025efficient}
\bibfield{author}{\bibinfo{person}{Jie Wu} {and} \bibinfo{person}{Mengshu Hou}.} \bibinfo{year}{2025}\natexlab{}.
\newblock \showarticletitle{An Efficient Retrieval-Based Method for Tabular Prediction with LLM}. In \bibinfo{booktitle}{\emph{Proceedings of the 31st International Conference on Computational Linguistics}}. \bibinfo{pages}{9917--9925}.
\newblock


\bibitem[Wu et~al\mbox{.}(2025)]%
        {wu2025tablebench}
\bibfield{author}{\bibinfo{person}{Xianjie Wu}, \bibinfo{person}{Jian Yang}, \bibinfo{person}{Linzheng Chai}, \bibinfo{person}{Ge Zhang}, \bibinfo{person}{Jiaheng Liu}, \bibinfo{person}{Xeron Du}, \bibinfo{person}{Di Liang}, \bibinfo{person}{Daixin Shu}, \bibinfo{person}{Xianfu Cheng}, \bibinfo{person}{Tianzhen Sun}, {et~al\mbox{.}}} \bibinfo{year}{2025}\natexlab{}.
\newblock \showarticletitle{Tablebench: A comprehensive and complex benchmark for table question answering}. In \bibinfo{booktitle}{\emph{Proceedings of the AAAI Conference on Artificial Intelligence}}, Vol.~\bibinfo{volume}{39}. \bibinfo{pages}{25497--25506}.
\newblock


\bibitem[Zha et~al\mbox{.}(2023)]%
        {zha2023tablegpt}
\bibfield{author}{\bibinfo{person}{Liangyu Zha}, \bibinfo{person}{Junlin Zhou}, \bibinfo{person}{Liyao Li}, \bibinfo{person}{Rui Wang}, \bibinfo{person}{Qingyi Huang}, \bibinfo{person}{Saisai Yang}, \bibinfo{person}{Jing Yuan}, \bibinfo{person}{Changbao Su}, \bibinfo{person}{Xiang Li}, \bibinfo{person}{Aofeng Su}, {et~al\mbox{.}}} \bibinfo{year}{2023}\natexlab{}.
\newblock \showarticletitle{Tablegpt: Towards unifying tables, nature language and commands into one gpt}.
\newblock \bibinfo{journal}{\emph{arXiv preprint arXiv:2307.08674}} (\bibinfo{year}{2023}).
\newblock


\bibitem[Zhan et~al\mbox{.}(2025)]%
        {zhan2025general}
\bibfield{author}{\bibinfo{person}{Marcus Zhan}, \bibinfo{person}{Kun Zhao}, \bibinfo{person}{Guodong Liu}, {and} \bibinfo{person}{Haoteng Tang}.} \bibinfo{year}{2025}\natexlab{}.
\newblock \showarticletitle{A General Paradigm for Fine-Tuning Large Language Models in Alzheimer’s Disease Diagnosis}. In \bibinfo{booktitle}{\emph{Proceedings of the AAAI Symposium Series}}, Vol.~\bibinfo{volume}{5}. \bibinfo{pages}{37--42}.
\newblock


\bibitem[Zhang et~al\mbox{.}(2024)]%
        {zhang2024vision}
\bibfield{author}{\bibinfo{person}{Jingyi Zhang}, \bibinfo{person}{Jiaxing Huang}, \bibinfo{person}{Sheng Jin}, {and} \bibinfo{person}{Shijian Lu}.} \bibinfo{year}{2024}\natexlab{}.
\newblock \showarticletitle{Vision-language models for vision tasks: A survey}.
\newblock \bibinfo{journal}{\emph{IEEE Transactions on Pattern Analysis and Machine Intelligence}} (\bibinfo{year}{2024}).
\newblock


\bibitem[Zhang et~al\mbox{.}(2023)]%
        {zhang2023generative}
\bibfield{author}{\bibinfo{person}{Tianping Zhang}, \bibinfo{person}{Shaowen Wang}, \bibinfo{person}{Shuicheng Yan}, \bibinfo{person}{Li Jian}, {and} \bibinfo{person}{Qian Liu}.} \bibinfo{year}{2023}\natexlab{}.
\newblock \showarticletitle{Generative Table Pre-training Empowers Models for Tabular Prediction}. In \bibinfo{booktitle}{\emph{Proceedings of the 2023 Conference on Empirical Methods in Natural Language Processing}}. \bibinfo{pages}{14836--14854}.
\newblock


\bibitem[Zhou et~al\mbox{.}(2023)]%
        {zhou2023algorithms}
\bibfield{author}{\bibinfo{person}{Hattie Zhou}, \bibinfo{person}{Arwen Bradley}, \bibinfo{person}{Etai Littwin}, \bibinfo{person}{Noam Razin}, \bibinfo{person}{Omid Saremi}, \bibinfo{person}{Josh Susskind}, \bibinfo{person}{Samy Bengio}, {and} \bibinfo{person}{Preetum Nakkiran}.} \bibinfo{year}{2023}\natexlab{}.
\newblock \showarticletitle{What algorithms can transformers learn? a study in length generalization}.
\newblock \bibinfo{journal}{\emph{arXiv preprint arXiv:2310.16028}} (\bibinfo{year}{2023}).
\newblock


\end{thebibliography}

\onecolumn
\appendix

\section*{Supplementary Materials}
\vspace{8pt}

\section{Prompt Formats for AD Prediction} 

Below, we provide example prompt–output pairs for each prompt format: zero-shot tabular, zero-shot serialized, few-shot tabular, and few-shot serialized. All patient data shown in these examples is entirely 
fabricated.

\subsubsection*{Zero-Shot Tabular}

\textbf{Prompt:}
\begin{lstlisting}
Below is an instruction that describes a task, paired with an input that provides further context. Write a response that appropriately completes the request.
### Instruction: Below is a table of patient records. Each column contains features related to Alzheimer's disease. Based on the information, predict whether the patient in the last row has Alzheimer's disease (1) or does not (0). Respond only with 1 or 0.
### Input:
AGE GENDER EDUCATION APOE4    FDG    AV45 CSF_ABETA(pg/ml) CSF_TAU(pg/ml) CSF_PTAU(pg/ml) WholeBrain Hippocampus Entorhinal Ventricles MidTemp Fusiform
 73 Female        11     0  1.131  1.4311           1163.2          305.2           29.48     978382        6728       3278      31733   21383    18321
### Response: 
\end{lstlisting}
\textbf{Expected Model Output:} 0

\subsubsection*{Zero-Shot Serialized}
\textbf{Prompt:}
\begin{lstlisting}
Below is an instruction that describes a task, paired with an input that provides further context. Write a response that appropriately completes the request.
### Instruction: Below is a table of patient records. Each column contains features related to Alzheimer's disease. Based on the information, predict whether the patient in the last row has Alzheimer's disease (1) or does not (0). Respond only with 1 or 0.
### Input:
A 73-year-old woman arrives for Alzheimer's Disease (AD) diagnosis. She has received 11 years of education, and she carries 0 copies of the APOE4 genetic variant. She received clinical examinations including cerebrospinal fluid (CSF) analysis, positron emission tomography (PET) imaging, and brain magnetic resonance imaging (MRI) imaging. For beta-amyloid pathology, she has an amyloid PET measure of 1.4311 and a CSF A-beta42 measure of 1163.2 pg/ml. For tau pathology, she has a CSF tau measure of 305.2 pg/ml and a CSF phosphorylated tau measure of 29.48 pg/ml. For MRI neuroimaging scans, she has whole brain volume of 978382, hippocampus region volume of 6728, entorhinal volume of 3278, ventricles volume of 31733, middle temporal lobe volume of 21383, and fusiform gyrus volume of 18321. In addition, her fluorodeoxyglucose (FDG) PET scan show an FDG measure of 1.131.
### Response: 
\end{lstlisting}

\textbf{Expected Model Output:} 0

\subsubsection*{Few-Shot Tabular}
 \textbf{Prompt:}
\begin{lstlisting}
Below is an instruction that describes a task, paired with an input that provides further context. Write a response that appropriately completes the request.
### Instruction: Below is a table of patient records. Each column contains features related to Alzheimer's disease. The last row is missing a value in the 'AlzheimersDisease' column. Based on the patterns in the other rows, predict whether the patient in the last row has Alzheimer's disease (1) or does not (0). Respond only with 1 or 0.
### Input:
 AGE GENDER  EDUCATION  APOE4    FDG   AV45  CSF_ABETA(pg/ml)  CSF_TAU(pg/ml)  CSF_PTAU(pg/ml)  WholeBrain  Hippocampus  Entorhinal  Ventricles  MidTemp  Fusiform AlzheimersDisease
  74 Female         12      1 1.4271 0.8999            1821.2           425.8            38.21     9052812         6788        3743       28326    26737     10474                 1
  64   Male         14      0 0.8377 0.9868            1555.0           168.2            12.44     1479873         8940        3546       26458    27277     25435                 0
  59 Female         20      1 1.9326 1.4429             637.0           280.2            18.33     1043476         9374        4344       36277    13452     15366                 0
  70 Female         18      2 1.6361 1.3294            1113.4           400.6            23.73     1135319         8747        3967       27978    15263     15436                 0
  73 Female         12      1 1.0631 1.0242            1800.0           502.2            48.13      988354         7252        3700       20665    16543     18425                 1
  60   Male         14      1 1.2367 1.1111             992.2           183.8            12.70     1242152         8630        4355       45372    35421     26453                 0
  75   Male         12      0 1.3745 1.8312             721.4           502.2            53.37     1006782         5834        3005       48327    14325     24352                 1
  73 Female          9      2 0.9327 1.8421            1004.2           487.6            48.42      893277         5032        2477       27652    15234     10043                 1
  73 Female         11      0 1.1313 1.4311            1163.2           305.2            29.48      978382         6728        3278       31733    21383     18321                 X
### Response: 
\end{lstlisting}

\textbf{Expected Model Output:} 0

\subsubsection*{Few-Shot Serialized }

\textbf{Prompt:}

\begin{lstlisting}[aboveskip=0.5\baselineskip]
Below is an instruction that describes a task, paired with an input that provides further context. Write a response that appropriately completes the request.
### Instruction: Below is a serialization of patient records. Each record contains features related to Alzheimer's disease. The last patient has a missing Alzheimer's diagnosis. Based on the patterns in the other records, predict whether the patient in the last record has Alzheimer's disease (1) or does not (0). Respond only with 1 or 0.
### Input:
A 74-year-old woman arrives for Alzheimer's Disease (AD) diagnosis and is diagnosed with Alzheimer's disease. She has received 12 years of education, and she carries 1 copy of the APOE4 genetic variant. She received clinical examinations including cerebrospinal fluid (CSF) analysis, positron emission tomography (PET) imaging, and brain magnetic resonance imaging (MRI) imaging. For beta-amyloid pathology, she has an amyloid PET measure of 0.8999 and a CSF A-beta42 measure of 1821.2 pg/ml. For tau pathology, she has a CSF tau measure of 425.8 pg/ml and a CSF phosphorylated tau measure of 38.21 pg/ml. For MRI neuroimaging scans, she has whole brain volume of 9052812, hippocampus region volume of 6788, entorhinal volume of 3743, ventricles volume of 28326, middle temporal lobe volume of 26737, and fusiform gyrus volume of 10474. In addition, her fluorodeoxyglucose (FDG) PET scan shows an FDG measure of 1.4271.
A 64-year-old man arrives for Alzheimer's Disease (AD) diagnosis and is not diagnosed with Alzheimer's disease. He has received 14 years of education, and he carries 0 copies of the APOE4 genetic variant. He received clinical examinations including cerebrospinal fluid (CSF) analysis, positron emission tomography (PET) imaging, and brain magnetic resonance imaging (MRI) imaging. For beta-amyloid pathology, he has an amyloid PET measure of 0.9868 and a CSF A-beta42 measure of 1555.0 pg/ml. For tau pathology, he has a CSF tau measure of 168.2 pg/ml and a CSF phosphorylated tau measure of 12.44 pg/ml. For MRI neuroimaging scans, he has whole brain volume of 1479873, hippocampus region volume of 8940, entorhinal volume of 3546, ventricles volume of 26458, middle temporal lobe volume of 27277, and fusiform gyrus volume of 25435. In addition, his fluorodeoxyglucose (FDG) PET scan shows an FDG measure of 0.8377.
A 59-year-old woman arrives for Alzheimer's Disease (AD) diagnosis and is not diagnosed with Alzheimer's disease. She has received 20 years of education, and she carries 1 copy of the APOE4 genetic variant. She received clinical examinations including cerebrospinal fluid (CSF) analysis, positron emission tomography (PET) imaging, and brain magnetic resonance imaging (MRI) imaging. For beta-amyloid pathology, she has an amyloid PET measure of 1.4429 and a CSF A-beta42 measure of 637.0 pg/ml. For tau pathology, she has a CSF tau measure of 280.2 pg/ml and a CSF phosphorylated tau measure of 18.33 pg/ml. For MRI neuroimaging scans, she has whole brain volume of 1043476, hippocampus region volume of 9374, entorhinal volume of 4344, ventricles volume of 36277, middle temporal lobe volume of 13452, and fusiform gyrus volume of 15366. In addition, her fluorodeoxyglucose (FDG) PET scan shows an FDG measure of 1.9326.
A 70-year-old woman arrives for Alzheimer's Disease (AD) diagnosis and is not diagnosed with Alzheimer's disease. She has received 18 years of education, and she carries 2 copies of the APOE4 genetic variant. She received clinical examinations including cerebrospinal fluid (CSF) analysis, positron emission tomography (PET) imaging, and brain magnetic resonance imaging (MRI) imaging. For beta-amyloid pathology, she has an amyloid PET measure of 1.3294 and a CSF A-beta42 measure of 1113.4 pg/ml. For tau pathology, she has a CSF tau measure of 400.6 pg/ml and a CSF phosphorylated tau measure of 23.73 pg/ml. For MRI neuroimaging scans, she has whole brain volume of 1135319, hippocampus region volume of 8747, entorhinal volume of 3967, ventricles volume of 27978, middle temporal lobe volume of 15263, and fusiform gyrus volume of 15436. In addition, her fluorodeoxyglucose (FDG) PET scan shows an FDG measure of 1.6361.
A 73-year-old woman arrives for Alzheimer's Disease (AD) diagnosis and is diagnosed with Alzheimer's disease. She has received 12 years of education, and she carries 1 copy of the APOE4 genetic variant. She received clinical examinations including cerebrospinal fluid (CSF) analysis, positron emission tomography (PET) imaging, and brain magnetic resonance imaging (MRI) imaging. For beta-amyloid pathology, she has an amyloid PET measure of 1.0242 and a CSF A-beta42 measure of 1800.0 pg/ml. For tau pathology, she has a CSF tau measure of 502.2 pg/ml and a CSF phosphorylated tau measure of 48.13 pg/ml. For MRI neuroimaging scans, she has whole brain volume of 988354, hippocampus region volume of 7252, entorhinal volume of 3700, ventricles volume of 20665, middle temporal lobe volume of 16543, and fusiform gyrus volume of 18425. In addition, her fluorodeoxyglucose (FDG) PET scan shows an FDG measure of 1.0631.
A 60-year-old man arrives for Alzheimer's Disease (AD) diagnosis and is not diagnosed with Alzheimer's disease. He has received 14 years of education, and he carries 1 copy of the APOE4 genetic variant. He received clinical examinations including cerebrospinal fluid (CSF) analysis, positron emission tomography (PET) imaging, and brain magnetic resonance imaging (MRI) imaging. For beta-amyloid pathology, he has an amyloid PET measure of 1.1111 and a CSF A-beta42 measure of 992.2 pg/ml. For tau pathology, he has a CSF tau measure of 183.8 pg/ml and a CSF phosphorylated tau measure of 12.70 pg/ml. For MRI neuroimaging scans, he has whole brain volume of 1242152, hippocampus region volume of 8630, entorhinal volume of 4355, ventricles volume of 45372, middle temporal lobe volume of 35421, and fusiform gyrus volume of 26453. In addition, his fluorodeoxyglucose (FDG) PET scan shows an FDG measure of 1.2367.
A 75-year-old man arrives for Alzheimer's Disease (AD) diagnosis and is diagnosed with Alzheimer's disease. He has received 12 years of education, and he carries 0 copies of the APOE4 genetic variant. He received clinical examinations including cerebrospinal fluid (CSF) analysis, positron emission tomography (PET) imaging, and brain magnetic resonance imaging (MRI) imaging. For beta-amyloid pathology, he has an amyloid PET measure of 1.8312 and a CSF A-beta42 measure of 721.4 pg/ml. For tau pathology, he has a CSF tau measure of 502.2 pg/ml and a CSF phosphorylated tau measure of 53.37 pg/ml. For MRI neuroimaging scans, he has whole brain volume of 1006782, hippocampus region volume of 5834, entorhinal volume of 3005, ventricles volume of 48327, middle temporal lobe volume of 14325, and fusiform gyrus volume of 24352. In addition, his fluorodeoxyglucose (FDG) PET scan shows an FDG measure of 1.3745.
A 73-year-old woman arrives for Alzheimer's Disease (AD) diagnosis, predict their diagnosis. She has received 9 years of education, and she carries 2 copies of the APOE4 genetic variant. She received clinical examinations including cerebrospinal fluid (CSF) analysis, positron emission tomography (PET) imaging, and brain magnetic resonance imaging (MRI) imaging. For beta-amyloid pathology, she has an amyloid PET measure of 1.8421 and a CSF A-beta42 measure of 1004.2 pg/ml. For tau pathology, she has a CSF tau measure of 487.6 pg/ml and a CSF phosphorylated tau measure of 48.42 pg/ml. For MRI neuroimaging scans, she has whole brain volume of 893277, hippocampus region volume of 5032, entorhinal volume of 2477, ventricles volume of 27652, middle temporal lobe volume of 15234, and fusiform gyrus volume of 10043. In addition, her fluorodeoxyglucose (FDG) PET scan shows an FDG measure of 0.9327.
### Response: 
\end{lstlisting}

\textbf{Expected Model Output:} 0

\section{Details of Reasoning Analysis}
Here, we present the reasoning prompt used in our interpretability analysis of TAP-GPT, designed with minimal CoT to limit confoundings, along with two representative examples (few-shot ICL k=8, 16) for tabular AD prediction.
All patient data shown in these examples is entirely fabricated.

\subsubsection*{Simple CoT Reasoning added to Few-Shot Tabular case.}
\textbf{Prompt:}
\begin{lstlisting}
Below is an instruction that describes a task, paired with an input that provides further context. Write a response that appropriately completes the request.
### Instruction: Below is a table of patient records. Each column contains features related to Alzheimer's disease. The last row is missing a value in the 'AlzheimersDisease' column. Based on the patterns in the other rows, predict whether the patient in the last row has Alzheimer's disease (1) or does not (0). Respond with your reasoning and the prediction answer (1 or 0).
### Input: <Same format as the Few-Shot Tabular case in Section A>
### Response: Let's think step-by-step
\end{lstlisting}

\subsubsection*{Example 1 Output with Reasoning (Correct)}
\textbf{K=8:}
\begin{lstlisting}
To predict whether the patient in the last row has Alzheimer's disease, we need to analyze the patterns in the other rows. 

1. **AGE**: The ages range from 61 to 79 years.
2. **GENDER**: Predominantly female, but there are also males.
3. **EDUCATION**: Education levels vary from 8 to 20 years.
4. **APOE4**: Some patients have APOE4 (1), while others do not (0).
5. **FDG**, **AV45**, **CSF_ABETA**, **CSF_TAU**, **CSF_PTAU**, **WholeBrain**, **Hippocampus**, **Entorhinal**, **Ventricles**, **MidTemp**, **Fusiform**: These values vary across patients, but generally, higher values for CSF_ABETA and lower values for CSF_TAU and CSF_PTAU are associated with lower risk of Alzheimer's disease.

Given the last row:
<Details ignored here>

**Reasoning**:
- The patient is male and has APOE4, which increases the risk.
- The CSF_ABETA level is low (740.4), and the CSF_TAU and CSF_PTAU levels are relatively high (332.0 and 31.58, respectively), indicating a higher risk.
- Other biomarkers like FDG and AV45 are within a range that might suggest some cognitive impairment but not necessarily Alzheimer's disease.

**Prediction**:
Based on the high risk factors (APOE4, low CSF_ABETA, high CSF_TAU and CSF_PTAU), it is likely that this patient has Alzheimer's disease.

**Answer: 1**
\end{lstlisting}

\textbf{True Label:} 1

\subsubsection*{Example 2 Output with Reasoning (Flawed)}
\textbf{K=16:}
\begin{lstlisting}

To predict whether the patient in the last row has Alzheimer's disease (1) or does not (0), we need to analyze the patterns in the other rows. Here are the steps:

1. **Age**: The age range is from 65 to 87 years.
2. **Gender**: Predominantly female, but there are also males.
3. **Education**: Varies from 10 to 20 years.
4. **APOE4**: Most patients have APOE4 (1), but some do not (0).
5. **FDG**: Values range from 0.9259 to 1.5015.
6. **AV45**: Values range from 1.0074 to 1.5009.
7. **CSF_ABETA**: Values range from 466.2 to 1701.0 pg/ml.
8. **CSF_TAU**: Values range from 10.85 to 421.6 pg/ml.
9. **CSF_PTAU**: Values range from 10.85 to 42.01 pg/ml.
<More details ignored here>

**Observations**:
- Patients with Alzheimer's disease (AlzheimersDisease = 1) tend to have higher values for CSF_ABETA, CSF_TAU, and CSF_PTAU.
- Patients without Alzheimer's disease (AlzheimersDisease = 0) tend to have lower values for these biomarkers.
- The presence of APOE4 (APOE4 = 1) is common among patients with Alzheimer's disease.

**Prediction**:
Given the last row's data:
- CSF_ABETA: 465.8 pg/ml (lower than many with Alzheimer's)
- CSF_TAU: 247.2 pg/ml (lower than many with Alzheimer's)
- CSF_PTAU: 25.07 pg/ml (lower than many with Alzheimer's)
- APOE4: 0 (not present)

Based on these observations, the patient in the last row is less likely to have Alzheimer's disease.

**Prediction Answer**: 0
\end{lstlisting}
\textbf{True Label:} 0

\end{document}